%% file: root.tex
\documentclass[10pt, conference]{IEEEtran} 

\IEEEoverridecommandlockouts                           

\usepackage{times}
\usepackage{graphicx} 
\usepackage{comment}
\usepackage{caption, subcaption}
\usepackage{booktabs}
\usepackage{array}
\usepackage{multirow}
\usepackage{multicol}
\usepackage{hyperref}

\usepackage{wrapfig}

\newcommand{\ferm}{CoDER }
\newcommand{\fermNoSpace}{CoDER}
\newcommand{\fermlongs}{CoDER }
\newcommand{\fermlong}{{\bf Co}ntrastive Pre-training and {\bf D}ata Augmentation for {\bf E}fficient {\bf R}obotic Learning (CoDER)}
\newcommand{\coder}{CoDER }

\input{math_commands}

\title{Learning Visual Robotic Control Efficiently with Contrastive Pre-training and Data Augmentation}

\author{%
Albert Zhan$^{*1}$, Ruihan (Philip) Zhao$^{*1}$,
Lerrel Pinto$^2$, Pieter Abbeel$^1$,
Michael Laskin$^1$%

\thanks{$^1$ University of California, Berkeley, $^2$ New York University
\texttt{\{albertzhan,philipzhao\}@berkeley.edu}}
\thanks{$^*$ Equal contribution}
}

\begin{document}
\vspace{3mm}
\maketitle
% \thispagestyle{empty}
% \pagestyle{empty}
% \IEEEpeerreviewmaketitle

%===============================================================================

\begin{abstract}
\input{content/abstract}

\end{abstract}

% \keywords{ Robot Learning, Reinforcement Learning, Unsupervised Pre-training, Data Augmentation} 

%===============================================================================

\input{content/introduction}

%===============================================================================

\input{content/background}

\input{figures/main_results_time}

\input{content/approach}

\input{content/results}

\input{content/related_work}
\input{content/limitations}

\input{content/acknowledgements.tex}
% \input{content/conclusion}

% The maximum paper length is 8 pages excluding references and acknowledgements, and 10 pages including references and acknowledgements

% \clearpage
% The acknowledgments are automatically included only in the final and preprint versions of the paper.
% \acknowledgments{If a paper is accepted, the final camera-ready version will (and probably should) include acknowledgments. All acknowledgments go at the end of the paper, including thanks to reviewers who gave useful comments, to colleagues who contributed to the ideas, and to funding agencies and corporate sponsors that provided financial support.}

%===============================================================================

\bibliographystyle{IEEEtran}
\bibliography{references}  % .bib

% \section{Appendix}
% \input{content/task_description}
% \input{content/baselines}
% \input{content/ablations_appendix}
\end{document}

%% file: math_commands.tex
\usepackage{amsmath,amsfonts,bm}

\def\eqref#1{equation~\ref{#1}}
\def\1{\bm{1}}

\DeclareMathAlphabet{\mathsfit}{\encodingdefault}{\sfdefault}{m}{sl}
\SetMathAlphabet{\mathsfit}{bold}{\encodingdefault}{\sfdefault}{bx}{n}

%% file: content/abstract.tex
Recent advances in unsupervised representation learning significantly improved the sample efficiency of training Reinforcement Learning policies in simulated environments. However, similar gains have not yet been seen for real-robot reinforcement learning. In this work, we focus on enabling data-efficient real-robot learning from pixels. We present 
{\bf Co}ntrastive Pre-training and {\bf D}ata Augmentation for {\bf E}fficient {\bf R}obotic Learning (CoDER),
% a \fermlong,
a method that utilizes data augmentation and unsupervised learning to achieve sample-efficient training of real-robot arm policies from sparse rewards. While contrastive pre-training, data augmentation, demonstrations, and reinforcement learning are alone insufficient for efficient learning, our main contribution is showing that the combination of these disparate techniques results in a simple yet data-efficient method. We show that, given only 10 demonstrations, a single robotic arm can learn sparse-reward manipulation policies from pixels, such as reaching, picking, moving, pulling a large object, flipping a switch, and opening a drawer in just 30 minutes of mean real-world training time.
We include videos and code on the project website \url{https://sites.google.com/view/efficient-robotic-manipulation/home}.

%%Data-efficient learning of manipulation policies from visual observations is an outstanding challenge for real-robot learning. While deep reinforcement learning (RL) algorithms have shown success learning policies from visual observations, they still require an impractical number of real-world data samples to learn effective policies. However, 

%Data-efficient learning of manipulation policies from visual observations is an outstanding challenge for real-robot learning. While deep reinforcement learning (RL) algorithms have shown success learning policies from visual observations, they still require an impractical number of real-world data samples to learn effective policies. However, recent advances in unsupervised representation learning and data augmentation significantly improved the sample efficiency of training RL policies on common simulated benchmarks. Building on these advances, we present a Framework for Efficient Robotic Manipulation (FERM) that utilizes data augmentation and unsupervised learning to achieve extremely sample-efficient training of robotic manipulation policies with sparse rewards. We show that, given only 10 demonstrations, a single robotic arm can learn sparse-reward manipulation policies from pixels, such as reaching, picking, moving, pulling a large object, flipping a switch, and opening a drawer in just 15-50 minutes of real-world training time.

%% file: content/introduction.tex
\section{Introduction}

Recent advances in deep reinforcement learning (RL) have given rise to unprecedented capabilities in autonomous decision making.  
Notable successes include learning to solve a diverse set of challenging video games \cite{mnih2015human,openai2019dota,agent57}, mastering complex classical games \cite{muzero}, and learning autonomous robotic control policies in both simulated 
% \cite{schulman2015trust,ppo,laskin_lee2020rad, hafner2019dream} 
\cite{laskin_lee2020rad, hafner2019dream} 
and real-world settings \cite{levine2015end,kalashnikov2018qt}. 
In particular, deep RL has been an effective method for learning diverse robotic manipulation policies such as grasping 
\cite{mahler2016dexnet} 
and dexterous in-hand manipulation of objects~\cite{andrychowicz2018learning}. 

 \begin{figure}%{0.5\textwidth}
  \begin{center}
    \includegraphics[width=0.4\textwidth]{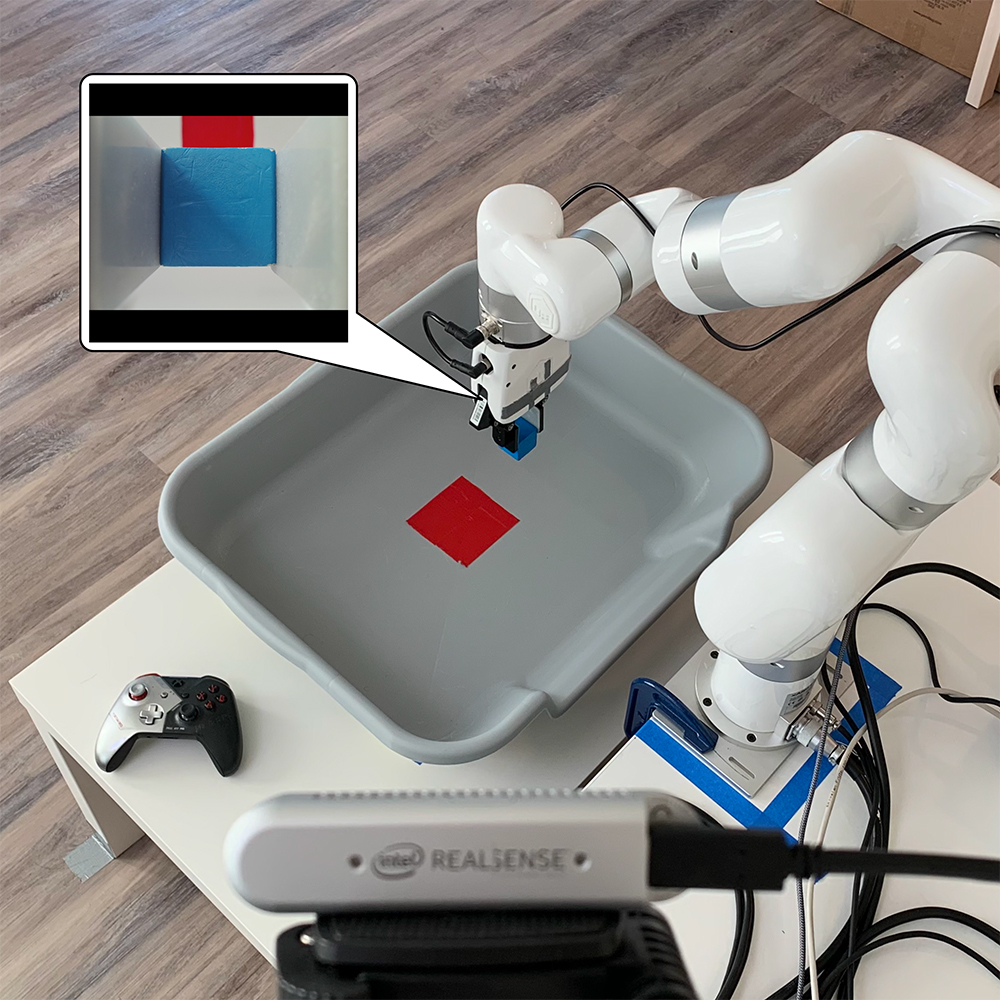}
  \end{center}
  \caption{
%   {\bf Co}ntrastive Pre-training and {\bf D}ata Augmentation for {\bf E}fficient {\bf R}obotic Learning (CoDER) 
    \fermlongs 
  enables robotic agents to learn skills directly from pixels in less than one hour of training. Our setup requires a robotic arm, two cameras, and a joystick to provide 10 demonstrations.
  }
\label{fig:teaser}
\vspace{-3mm}
\end{figure}
 
 However, to date, general purpose RL algorithms have been extremely sample inefficient, which has limited their widespread adoption in the field of robotics. State-of-the-art RL algorithms for discrete  \cite{hessel2017rainbow} and continuous  \cite{lillicrap2015continuous} control often require tens of millions of environment interactions to learn effective policies from image input \cite{tassa2018deepmind}, while training the Dota5 agent \cite{openai2019dota} to perform competitively to human experts required an estimated 180 human-years of game play. 
 Even when the underlying proprioceptive state is accessible, sparse reward robotic manipulation still needs millions of training samples \cite{andrychowicz2017her}, to achieve reliable success rates on fundamental tasks such as reaching, picking, pushing, and placing objects. 

Another common approach to learned robotic control is through imitation learning
\cite{zhang2018DeepIL,young2020, florence2022implicit}, 
% \cite{zhang2018DeepIL,ho2016gail,duan2017osil,finn17osil,young2020, florence2022implicit},
where a large number of expert demonstrations are collected and the policy is extracted through supervised learning by regressing onto the expert trajectories. 
However, imitation learning can take up to hundreds or thousands of expert demonstrations \cite{transporter, selfsupervisedcorrespondance, florence2022implicit}, which are laborious to collect, and the resulting policies are bounded by the quality of expert demonstrations. 
It would be more desirable to learn the optimal policy required to solve a particular task autonomously. 

In this work, rather than relying on transferring policies from simulation or labor intensive human input through imitation learning or environment engineering, 
we investigate how pixel-based RL applied to real robots can be made data-efficient.
Recent progress in unsupervised representation learning \cite{laskin2020curl,stooke2020atc} and data augmentation 
\cite{laskin_lee2020rad, kostrikov2020image} has significantly improved the efficiency of learning with RL in simulated robotic \cite{tassa2018deepmind} and video game \cite{bellemare2013arcade} environments. 
The primary strength of these methods is learning high quality representations from image input either explicitly through unsupervised learning or implicitly via data augmentation.

\input{figures/figure_architecture}

Building on these advances, we propose 
\fermlong, shown in Figure~\ref{fig:architecture}.
\ferm utilizes off-policy RL with data augmentation along with unsupervised pre-training to learn efficiently with a simple three-staged procedure. First, a small number of (10) demonstrations are collected and stored in a replay buffer. Second, the convolutional encoder weights are initialized with unsupervised contrastive pre-training on the demonstration data. Third, an off-policy RL algorithm is trained with augmented images on both data collected online during training and the initial demonstrations. Our core contribution is the novel combination of contrastive pre-training, online data augmentations, and utilizing a small number of demonstrations that together enable  efficient real-robot learning from pixels. In contrast, prior leading algorithms that utilize these components individually are unable to learn efficiently on real robots.

We summarize our key contributions and benefits of the \ferm algorithm:
{\it (1) Data-efficiency:} As shown in Figure~\ref{fig:res_time_bp}, \ferm enables learning optimal policies on 6 diverse manipulation tasks, {\bf in 15-50 minutes of total training time} for each task. These tasks include reaching, pushing, moving, pulling a large object, flipping a switch and drawer opening, as shown in Figure~\ref{fig:envs}.
{\it (2) Real-robot deployment:} \ferm trains efficiently on real robotic hardware.
% while prior related approaches that were successful in simulation~\cite{laskin2020curl,laskin_lee2020rad} fail to learn robust real-robot policies. 
{\it (3) Simplicity:} \ferm is a 
% novel 
combination of existing ideas such as contrastive pre-training, data augmentation, and demonstrations that results in a simple and easy to reproduce algorithm. {\it (4) General \& lightweight setup:} Our setup requires a robot, one GPU, two RGB cameras, a handful of demonstrations, and a sparse reward function. 
These requirements are quite lightweight relative to setups that rely on Sim2Real, motion capture, multiple robots, point-cloud estimates, or engineering dense rewards. 
% To the best of our knowledge, t
This work 
% is the first to 
shows how recent advances in contrastive learning and data augmentation can enable efficient real-robot reinforcement learning from pixels.

%% file: figures/figure_architecture.tex
\begin{figure*}[!ht!]
    \vspace{5pt}
    \includegraphics[width = \textwidth]{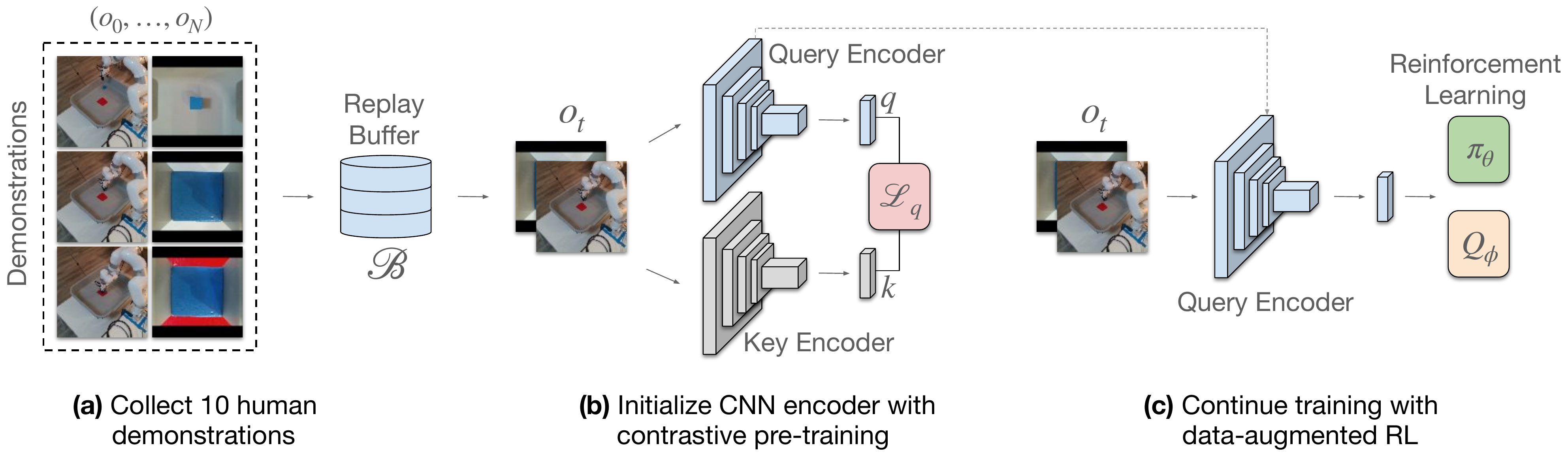} 
  \caption{The \coder architecture. 
  \textbf{(a)} Demonstrations are collected, and stored in a replay buffer. \textbf{(b)} The observations from the demonstrations are used to pre-train the encoder with a
  contrastive loss. 
  \textbf{(c)} The encoder and replay buffer are then used to train an RL agent using an off-policy data-augmented RL algorithm.}
\label{fig:architecture}
\vspace{-3mm}
\end{figure*}

%% file: content/background.tex
\section{Background}

{\bf Soft Actor Critic:}
The Soft Actor Critic (SAC) \cite{haarnoja2018soft} is an off-policy RL algorithm that jointly learns an action-conditioned state value function through Q learning and a stochastic policy by maximizing expected returns. 
SAC is a state-of-the-art model-free RL algorithm for continuous control from state \cite{haarnoja2018soft} and, in the presence of data augmentations, from pixels as well \cite{laskin_lee2020rad, kostrikov2020image}. 
In simulated benchmarks, such as DeepMind control \cite{tassa2018deepmind}, SAC is as data-efficient from pixels as it is from state \cite{laskin_lee2020rad}. 
For this reason, we utilize it as our base RL algorithm for sparse-reward manipulation in this work. As an actor-critic method, SAC learns an actor policy $\pi_\theta$  and an ensemble of critics $Q_{\phi_1}$ and $Q_{\phi_2}$. 

To learn the actor policy, samples are collected stochastically from $\pi_\theta$ such that $
a_\theta (o,\xi) \sim \tanh \left (\mu_\theta (o) + \sigma_\theta (o) \odot \xi \right )$, where $\xi \sim \mathcal N (0,I)$ is a sample from a normalized Gaussian noise vector, and then trained to maximize the expected return as shown in eq. \ref{eq:actorloss}.

\input{figures/figure_environments}

\begin{equation}\label{eq:actorloss}
   \mathcal L (\theta) = \mathbb{E}_{a \sim \pi} \left [ Q^\pi (o,a) - \alpha \log \pi_\theta (a|o) \right ]
\end{equation}

Simultaneously to learning the policy, SAC also trains the critics $Q_{\phi_1}$ and $Q_{\phi_2}$ to minimize the Bellman equation in \ref{eq:qmsbe}. Here, a transition  $t = (o,a,o',r,d)$ is sampled from the replay buffer $\mathcal{B}$, where $(o,o')$ are consecutive timestep observations, $a$ is the action, $r$ is the reward, and $d$ is the terminal flag.

\begin{equation}\label{eq:qmsbe}
   \mathcal L (\phi_{i},\mathcal{B}) = \mathbb{E}_{t \sim \mathcal B} \left [\left ( Q_{\phi_i}(o,a) - \left( r+\gamma(1-d) Q_{\text{targ}}  \right )\right )^2 \right]
\end{equation}

The function $Q_{\text{targ}}$ is the target value that the critics are trained to match, defined in \ref{eq:target}. The target is the entropy regularized exponential moving average (EMA) of the critic ensemble parameters, which we denote as $\bar Q_\phi$.

\begin{equation}\label{eq:target}
   Q_{\text{targ}}  = \left (\min_{i=1,2} \bar Q_{\phi_i} (o',a') - \alpha \log \pi_\theta(a'|o')\right )
\end{equation}

where $(a',o')$ are the consecutive timestep action and observation, and $\alpha$ is a positive action-entropy coefficient. A non-zero action-entropy term improves exploration -- the higher the value of $\alpha$ the more entropy maximization is prioritized over optimizing the value function.

%Contrastive learning \cite{hadsell2006dimensionality, lecun2006tutorial, oord2018representation, wu2018unsupervised, he2019momentum} is a paradigm for unsupervised representation learning that aims to maximize agreement between similar pairs of data while minimizing it between dissimilar ones. This type of representation learning has seen a recent resurgence in the field of computer vision where it was shown \cite{chen2020simclr,kaiming2019moco,henaff2019data}
%that representations pre-trained with a contrastive loss on a corpus of unlabeled ImageNet data, are effective for downstream classification tasks, matching and sometimes outperforming fully supervised learning and significantly outperforming it when the percentage of available labels per data point is small.  

{\bf Unsupervised Contrastive Pretraining:} Contrastive learning \cite{hadsell2006dimensionality, oord2018representation, wu2018unsupervised, he2019momentum,chen2020simclr,kaiming2019moco,henaff2019data} aims to maximize agreement between positive examples in data while minimizing agreement between negative examples. Contrastive methods require the specification of {\it query-key} pairs, also known as {\it anchors} and {\it positives}, which are similar data pairs whose agreement needs to be maximized. Given a query $q$
%$q \in \mathbb Q = \{q_0, q_1, \dots \} $ 
and a key $k$
%$k \in \mathbb K= \{k_0, k_1, \dots \}$
, we seek to maximize the score $f_{\text{score}}(q,k)$ between them while minimizing them between the query $q$ and negative examples in the dataset $k_-$. The score function is most often represented as an inner product, such as a dot product $f_{\text{score}}(q,k) = q^T k$ \cite{wu2018unsupervised, he2019momentum} or a bilinear product $f_{\text{score}}(q,k) = q^TWk$ \cite{oord2018representation, henaff2019data}, while other Euclidean metrics are also available \cite{schroff2015facenet, wang2015unsupervised}. Modern contrastive approaches \cite{chen2020simclr,kaiming2019moco,henaff2019data,laskin2020curl} employ the InfoNCE loss \cite{oord2018representation}, which is described in \ref{eq:infonce} and can also be interpreted as a multi-class cross entropy classification loss with $K$ classes. 

%Since the specification of positive query-key pairs is a design choice, it is usually straightforward to extract such pairs from the unlabeled dataset of interest. 
%However, the exact extraction of negatives can be challenging without prior knowledge due to the lack of labels. 
%For this reason, contrastive methods usually approximate negative sampling with Noise Contrastive Estimation (NCE) \cite{nce}, which effectively generates negatives by sampling noisily from the dataset. 
%\input{figures/main_results_barplot}
%In particular, modern contrastive approaches \cite{chen2020simclr,kaiming2019moco,henaff2019data,laskin2020curl} employ the InfoNCE loss \cite{oord2018representation}, which is described in Equation \ref{eq:infonce} and can also be interpreted as a multi-class cross entropy classification loss with $K$ classes. 

\begin{equation}\label{eq:infonce}
   \mathcal L_{q} = \log \frac{\exp(q^T W k)}{ \exp \left (\sum_{i=0}^{K} \exp(q^T W k_{i}) \right )}
\end{equation}

In the computer vision setting, a simple and natural choice of query-key specification is to define queries and keys as two data augmentations of the same image. 
This approach, called instance discrimination, is used in most of the state-of-the-art representation learning methods for static images \cite{chen2020simclr,kaiming2019moco} as well as RL from pixels \cite{laskin2020curl}. 
In the minibatch setting, which we also employ in this work, the InfoNCE loss is computed by sampling $K = \{x_1,\dots,x_K\}$ images from the dataset, generating queries $Q = \{q_1,\dots,q_K\}$ and keys $K = \{k_1 ,\dots,k_K \}$ with stochastic data augmentations $q_i, k_i = \text{aug}(x_i)$, and using each augmented datapoint $x_i$ as positives while the rest of the images are negatives.

%% file: figures/figure_environments.tex
\newcommand{\singlewidth}{.050}
\newcommand{\doublewidth}{.1}
\newcommand{\triplewidth}{.150}
\begin{figure*}[!ht]
    % \vspace{5pt}
    \centering
    % \begin{subfigure}[c]{0.03\linewidth}
    %   \includegraphics[width=.8\linewidth]{figures/arrow.png}
    %   \vspace{0.5cm}
    % \end{subfigure}
    % \hspace{.005\linewidth}
    \begin{subfigure}{\triplewidth\linewidth}
      \begin{subfigure}[c]{.64\linewidth}
        \centering
        \includegraphics[width=\linewidth, trim={76px 0px 92px 0px}, clip]{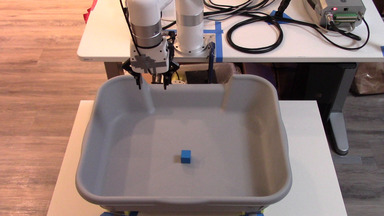}\\
        \includegraphics[width=\linewidth, trim={76px 0px 92px 0px}, clip]{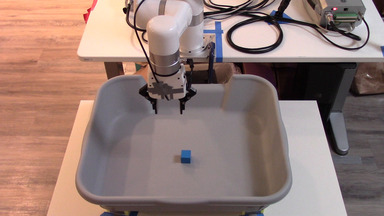}\\
        \includegraphics[width=\linewidth, trim={76px 0px 92px 0px}, clip]{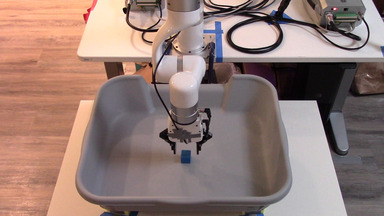}
      \end{subfigure}
      \begin{subfigure}[c]{.311\linewidth}
        \centering
        \includegraphics[width=\linewidth]{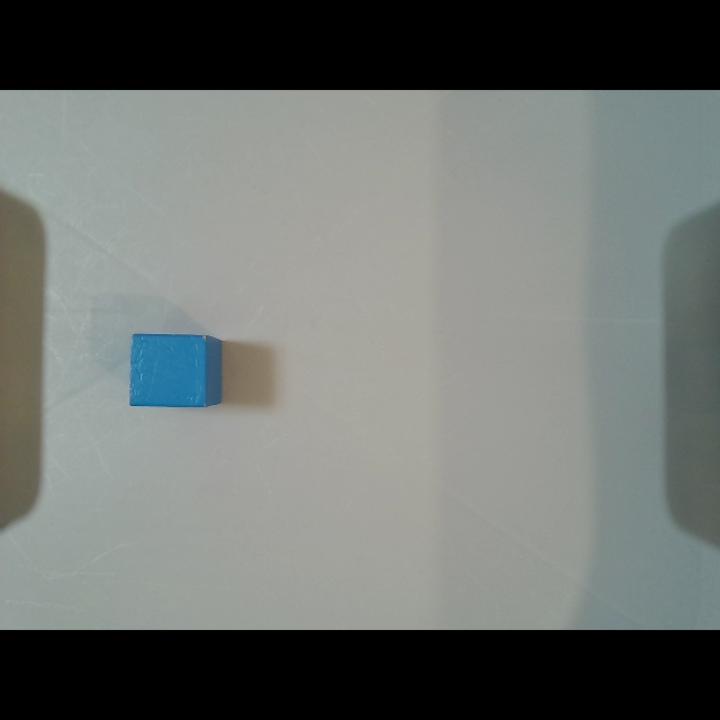}\\
        \includegraphics[width=\linewidth]{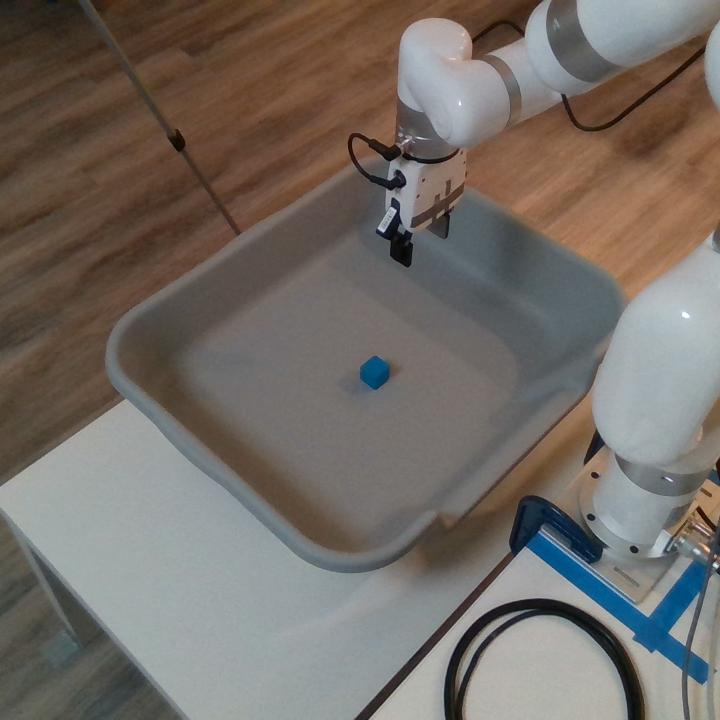}\\
        \includegraphics[width=\linewidth]{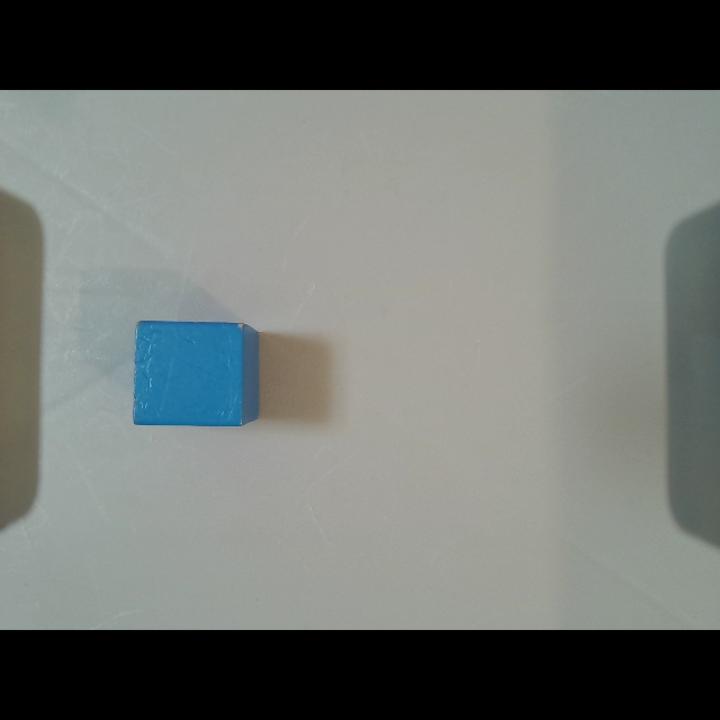}\\
        \includegraphics[width=\linewidth]{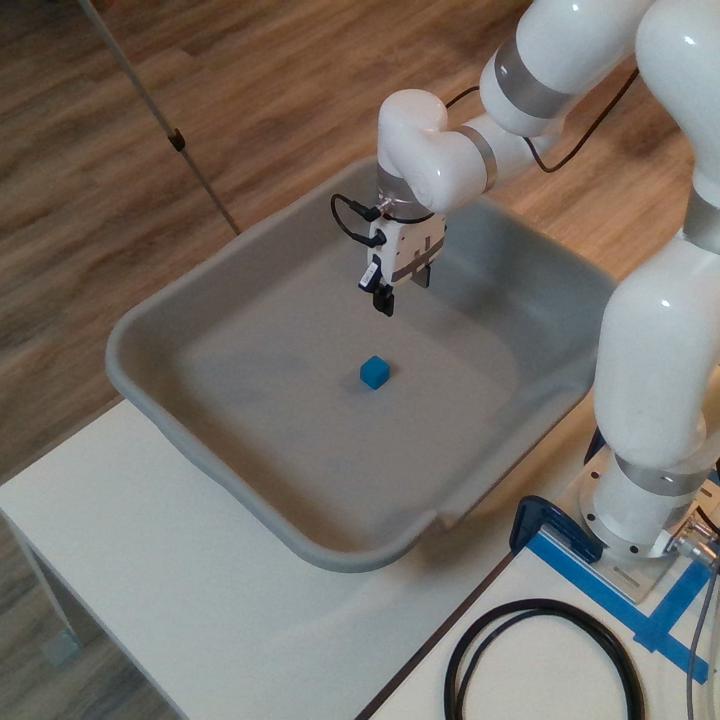}\\
        \includegraphics[width=\linewidth]{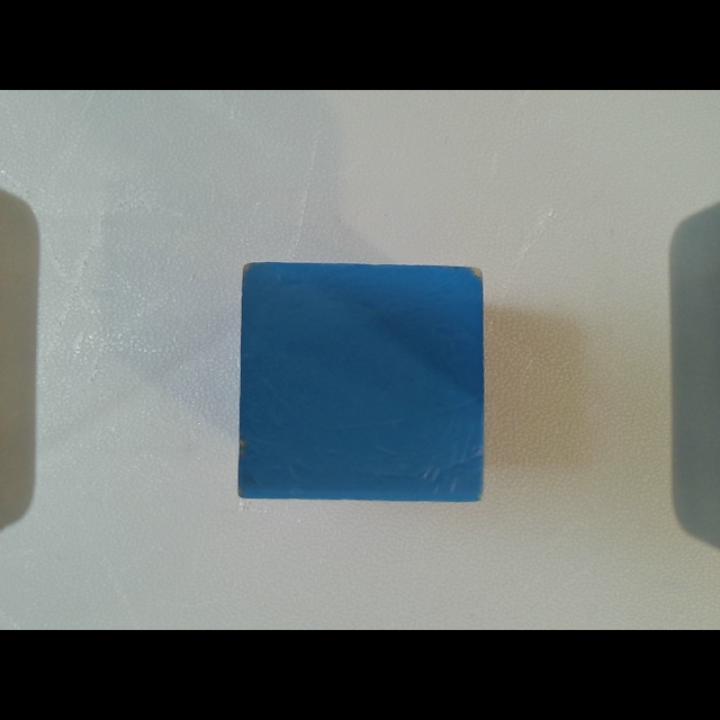}\\
        \includegraphics[width=\linewidth]{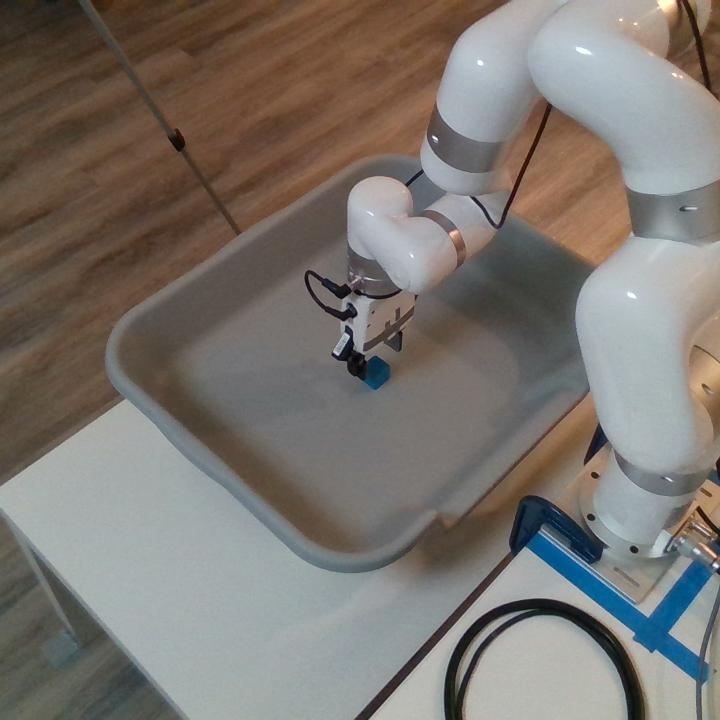}
      \end{subfigure}
      \caption{Reach}
    \end{subfigure}
    \hspace{.005\linewidth}
    \begin{subfigure}{\triplewidth\linewidth}
      \begin{subfigure}[c]{.64\linewidth}
        \centering
        \includegraphics[width=\linewidth, trim={76px 0px 92px 0px}, clip]{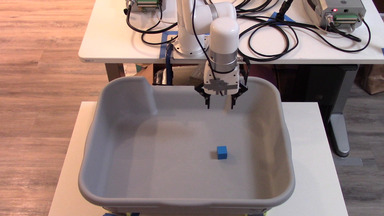}\\
        \includegraphics[width=\linewidth, trim={76px 0px 92px 0px}, clip]{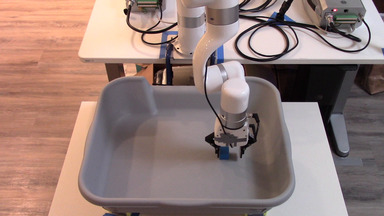}\\
        \includegraphics[width=\linewidth, trim={76px 0px 92px 0px}, clip]{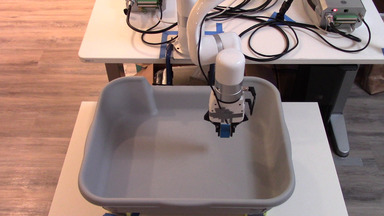}
      \end{subfigure}
      \begin{subfigure}[c]{.311\linewidth}
        \centering
        \includegraphics[width=\linewidth]{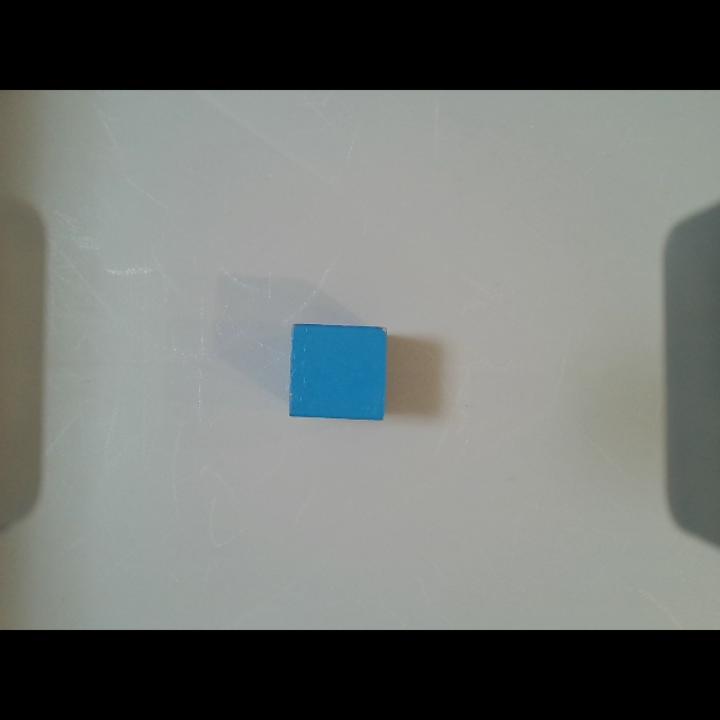}\\
        \includegraphics[width=\linewidth]{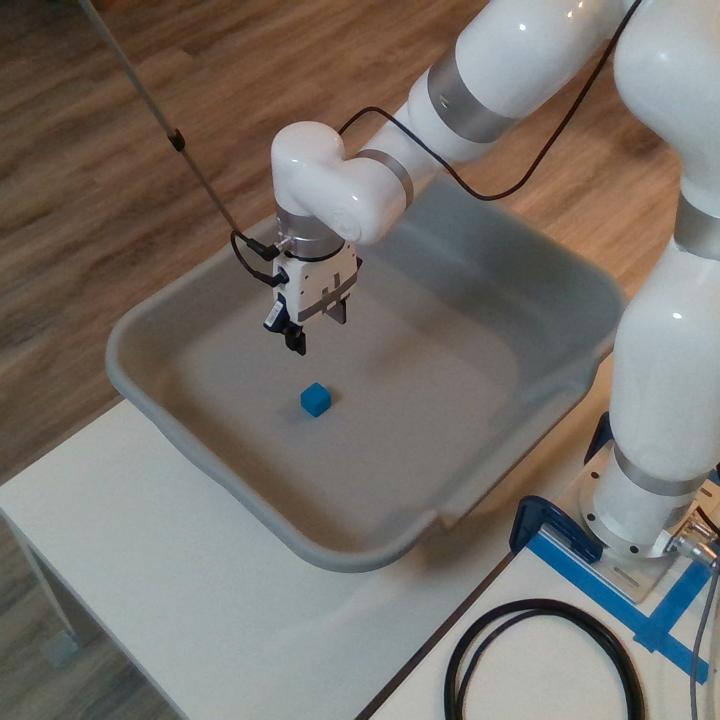}\\
        \includegraphics[width=\linewidth]{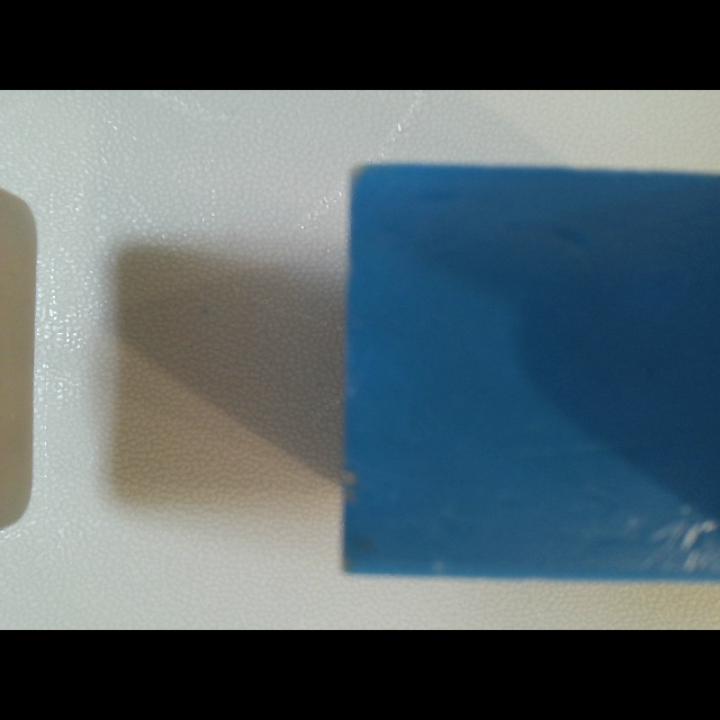}\\
        \includegraphics[width=\linewidth]{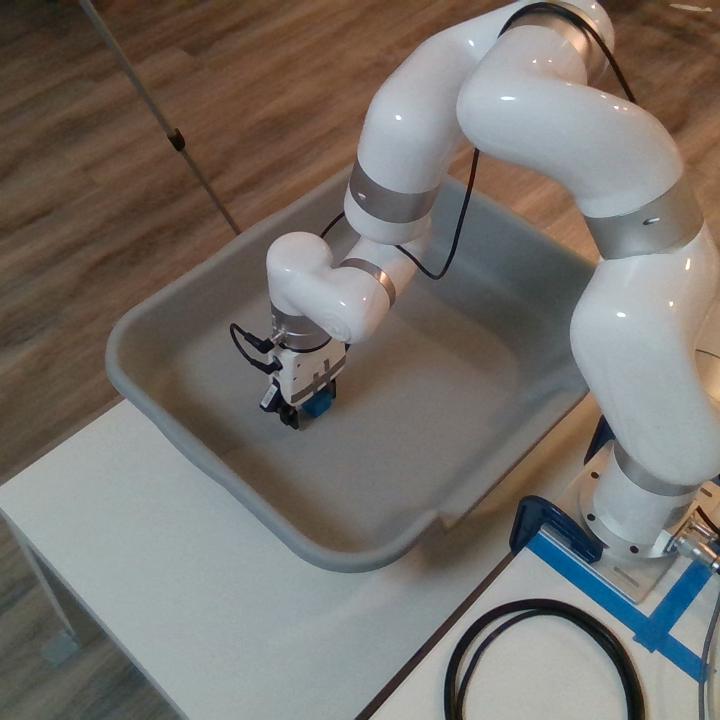}\\
        \includegraphics[width=\linewidth]{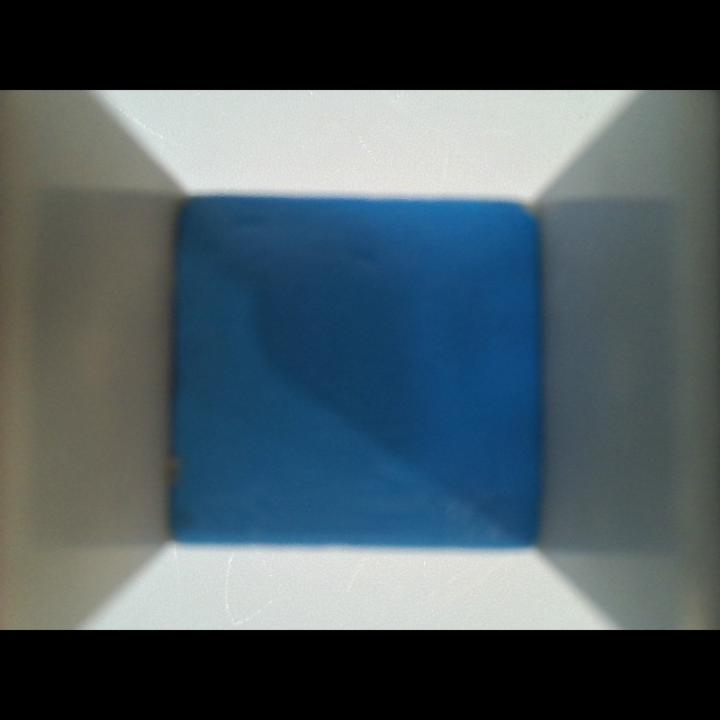}\\
        \includegraphics[width=\linewidth]{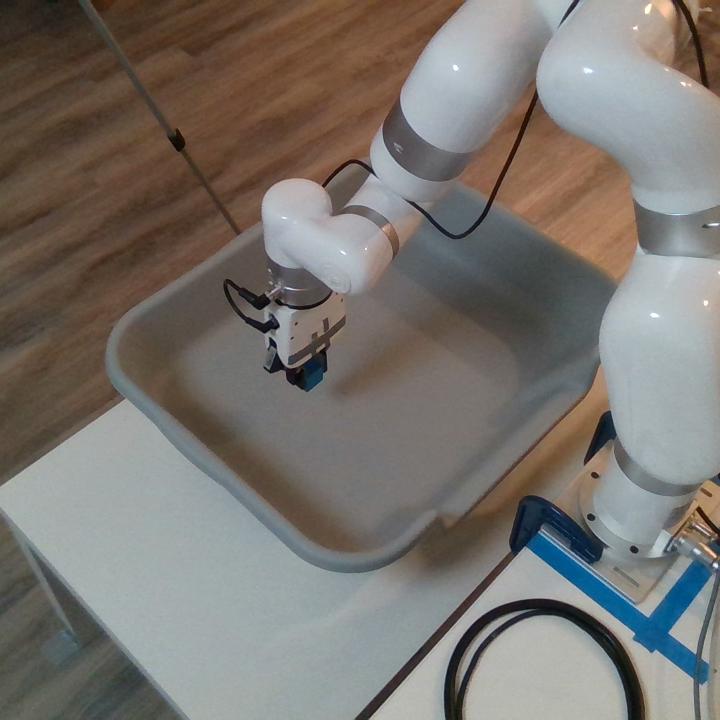}
      \end{subfigure}
      \caption{Pickup}
    \end{subfigure}
    \hspace{.005\linewidth}
    \begin{subfigure}{\triplewidth\linewidth}
      \begin{subfigure}[c]{.64\linewidth}
        \centering
        \includegraphics[width=\linewidth, trim={76px 0px 92px 0px}, clip]{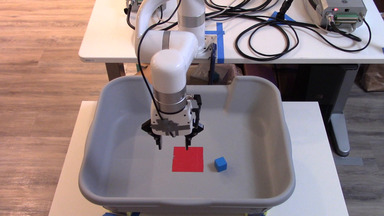}\\
        \includegraphics[width=\linewidth, trim={76px 0px 92px 0px}, clip]{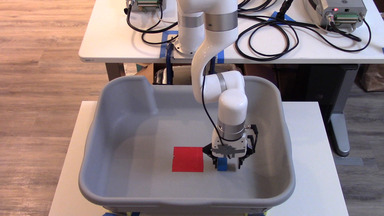}\\
        \includegraphics[width=\linewidth, trim={76px 0px 92px 0px}, clip]{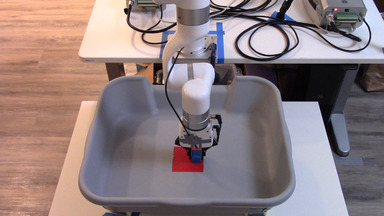}
      \end{subfigure}
      \begin{subfigure}[c]{.311\linewidth}
        \centering
        \includegraphics[width=\linewidth]{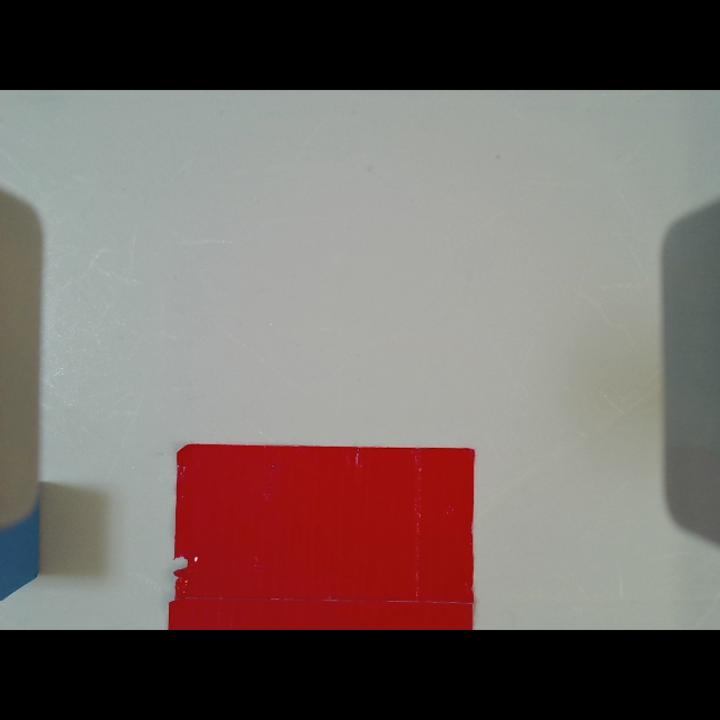}\\
        \includegraphics[width=\linewidth]{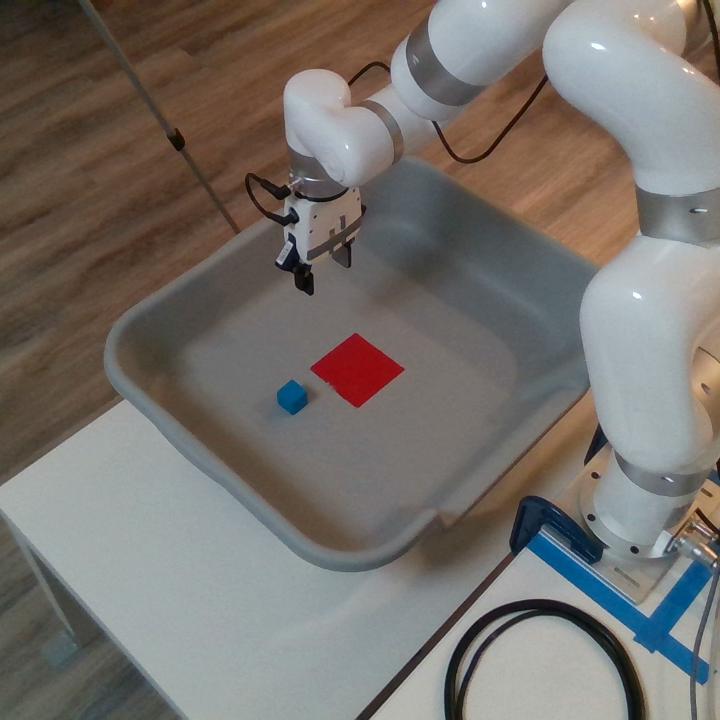}\\
        \includegraphics[width=\linewidth]{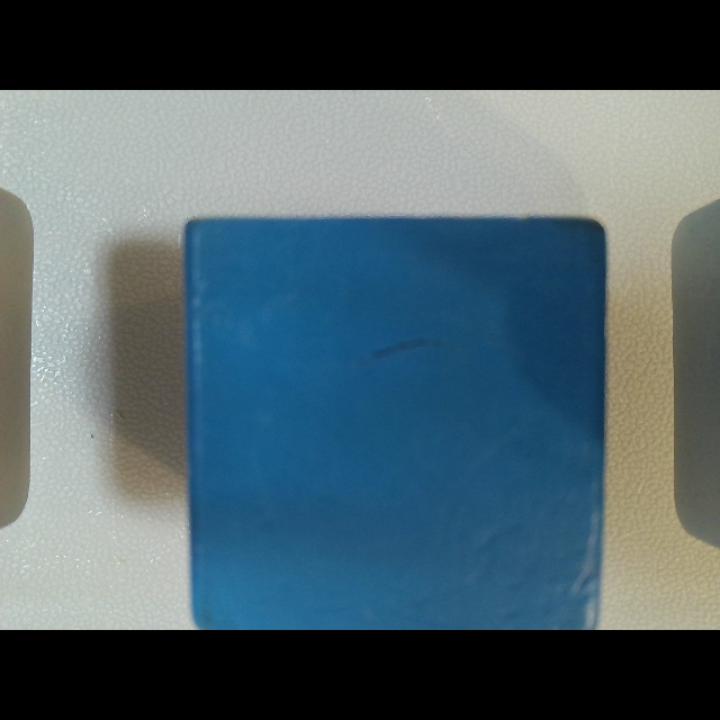}\\
        \includegraphics[width=\linewidth]{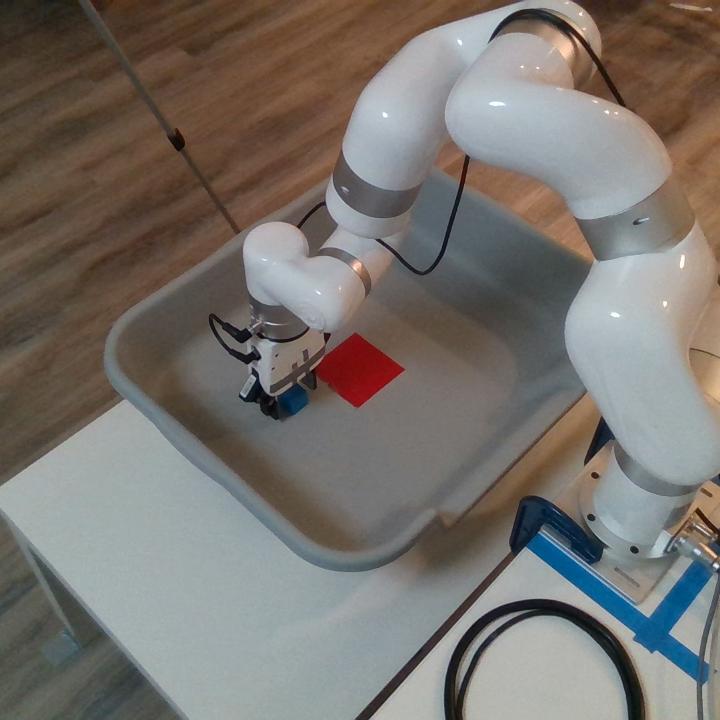}\\
        \includegraphics[width=\linewidth]{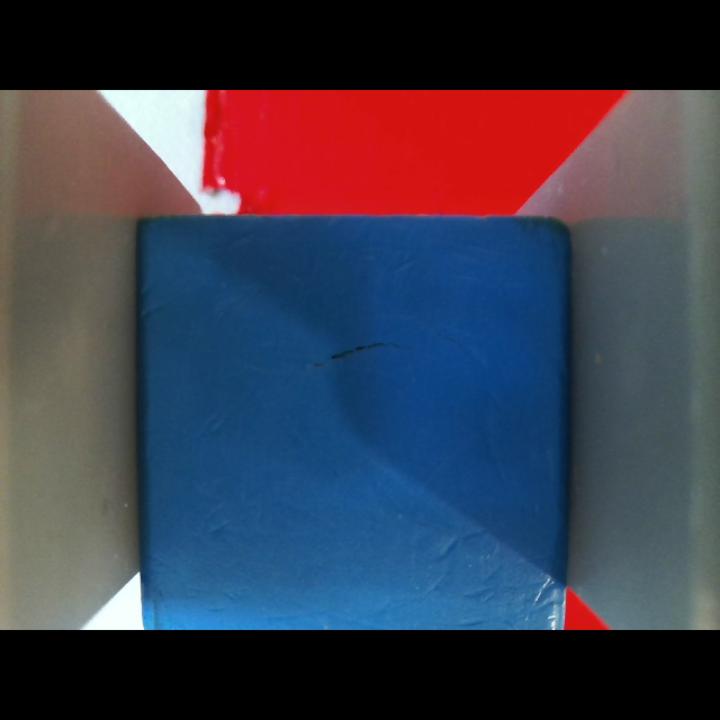}\\
        \includegraphics[width=\linewidth]{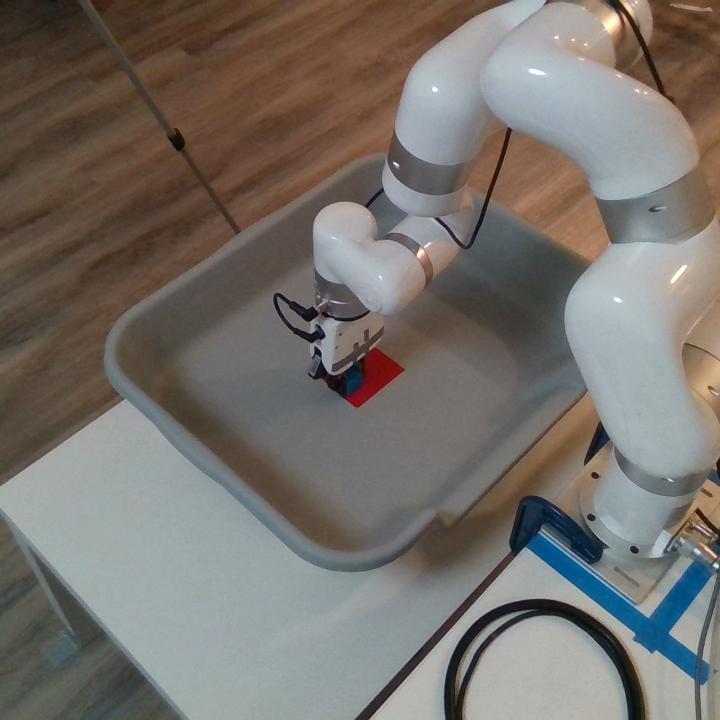}
      \end{subfigure}
      \caption{Move}
    \end{subfigure}
    \hspace{.005\linewidth}
    \begin{subfigure}{\triplewidth\linewidth}
      \begin{subfigure}[c]{.64\linewidth}
        \centering
        \includegraphics[width=\linewidth, trim={76px 0px 92px 0px}, clip]{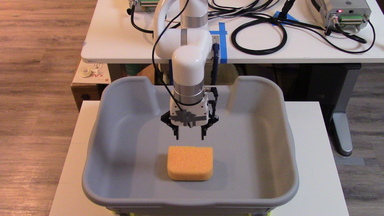}\\
        \includegraphics[width=\linewidth, trim={76px 0px 92px 0px}, clip]{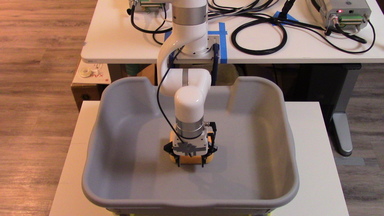}\\
        \includegraphics[width=\linewidth, trim={76px 0px 92px 0px}, clip]{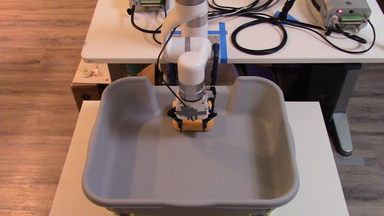}
      \end{subfigure}
      \begin{subfigure}[c]{.311\linewidth}
        \centering
        \includegraphics[width=\linewidth]{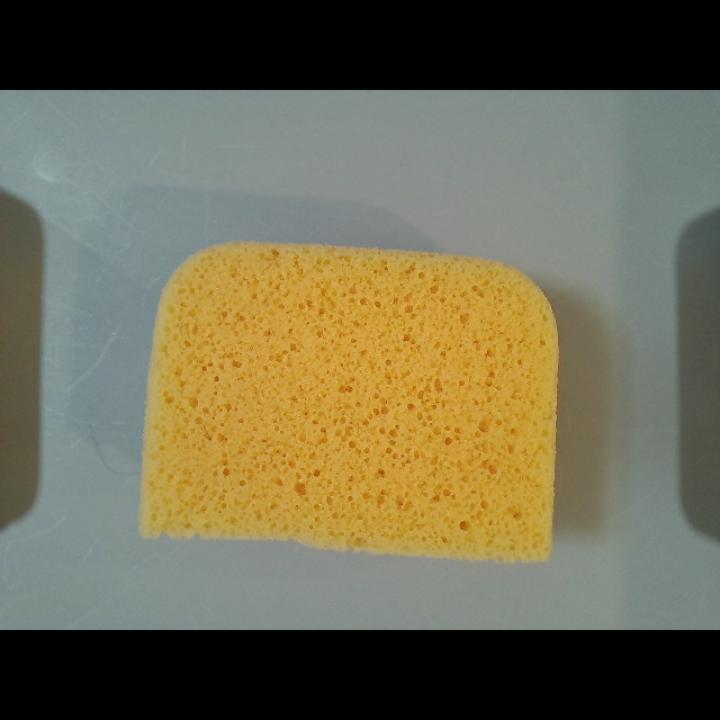}\\
        \includegraphics[width=\linewidth]{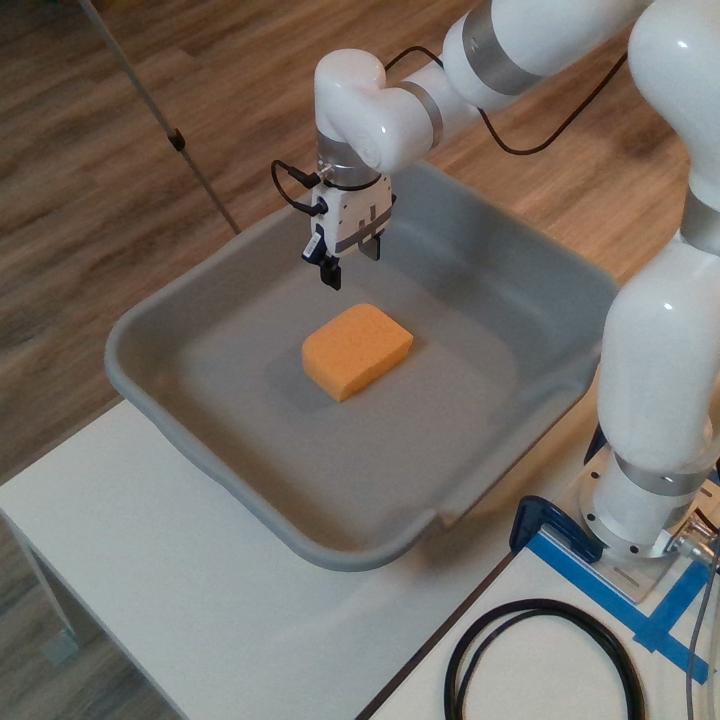}\\
        \includegraphics[width=\linewidth]{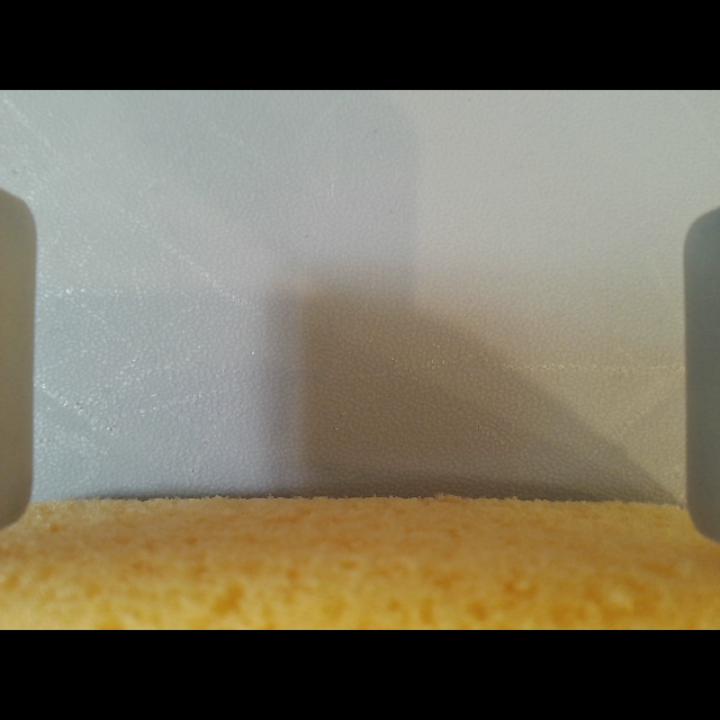}\\
        \includegraphics[width=\linewidth]{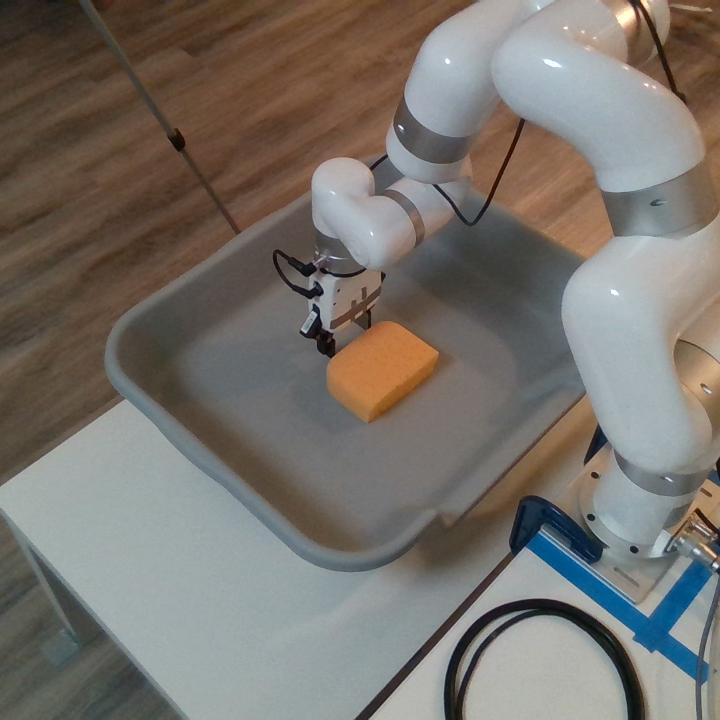}\\
        \includegraphics[width=\linewidth]{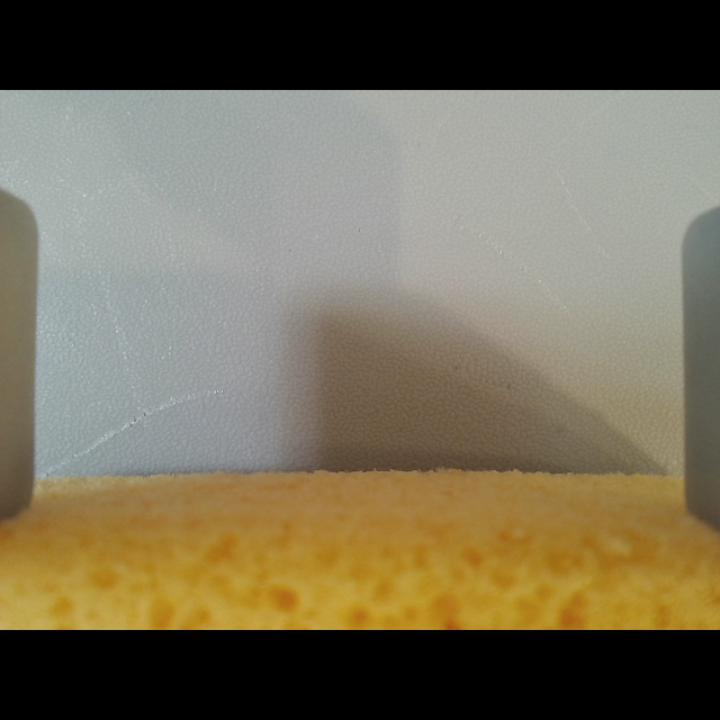}\\
        \includegraphics[width=\linewidth]{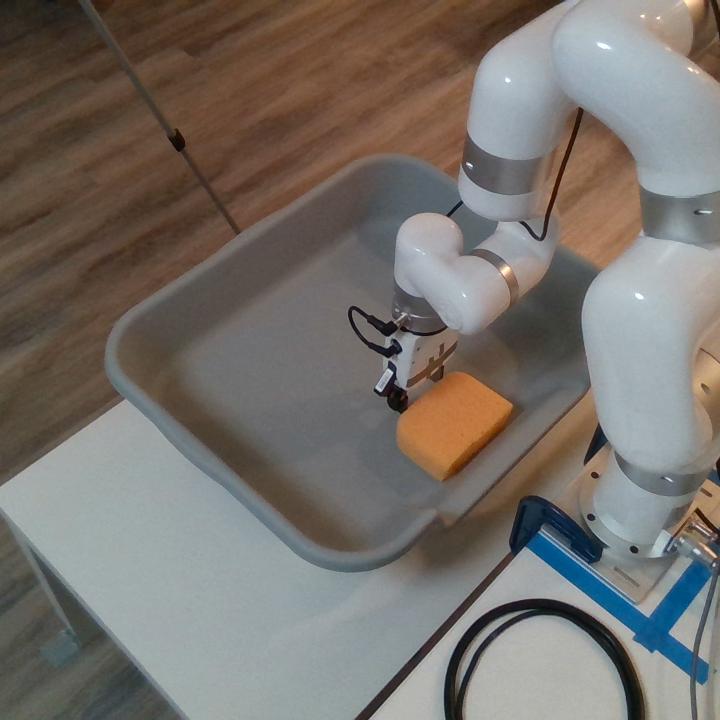}
      \end{subfigure}
      \caption{Pull}
    \end{subfigure}
    \hspace{.005\linewidth}
    \begin{subfigure}{\triplewidth\linewidth}
      \begin{subfigure}[c]{.64\linewidth}
        \centering
        \includegraphics[width=\linewidth, trim={76px 0px 92px 0px}, clip]{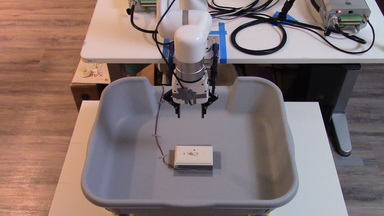}\\
        \includegraphics[width=\linewidth, trim={76px 0px 92px 0px}, clip]{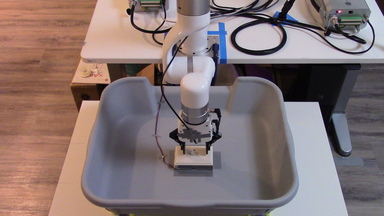}\\
        \includegraphics[width=\linewidth, trim={76px 0px 92px 0px}, clip]{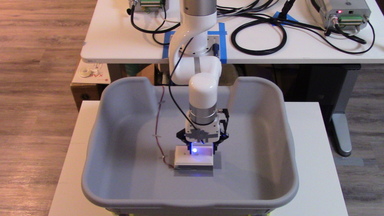}
      \end{subfigure}
      \begin{subfigure}[c]{.311\linewidth}
        \centering
        \includegraphics[width=\linewidth]{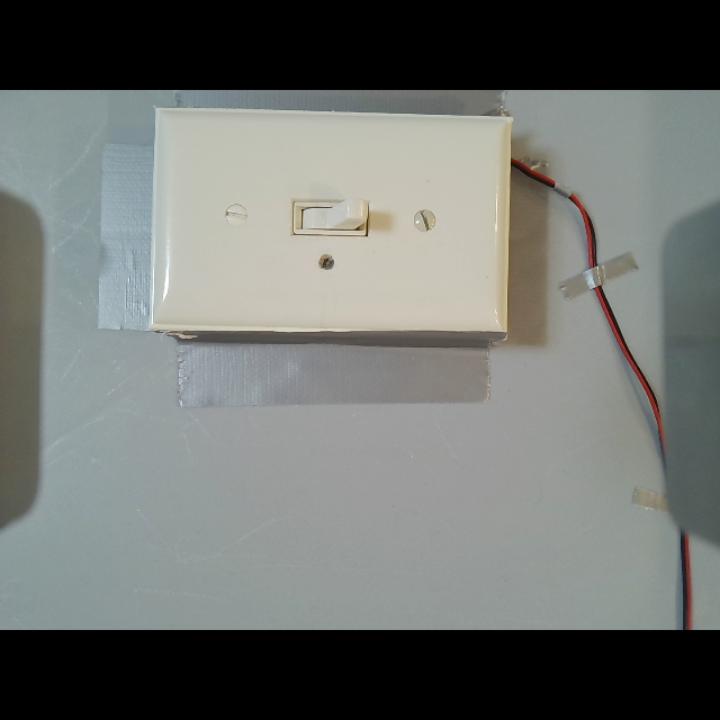}\\
        \includegraphics[width=\linewidth]{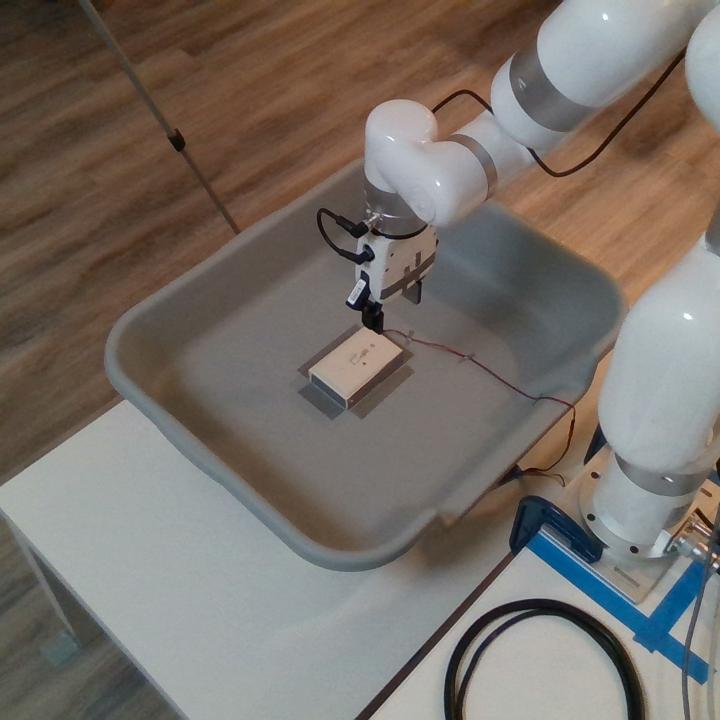}\\
        \includegraphics[width=\linewidth]{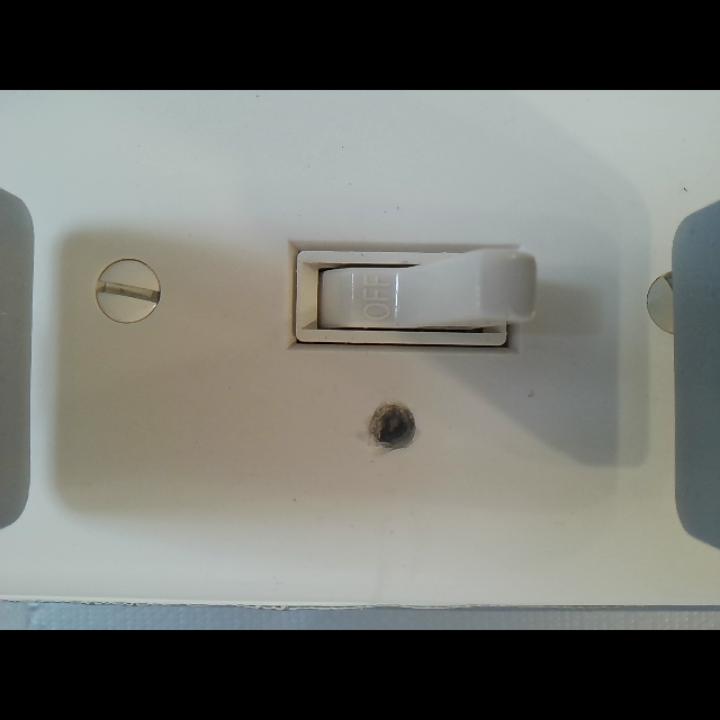}\\
        \includegraphics[width=\linewidth]{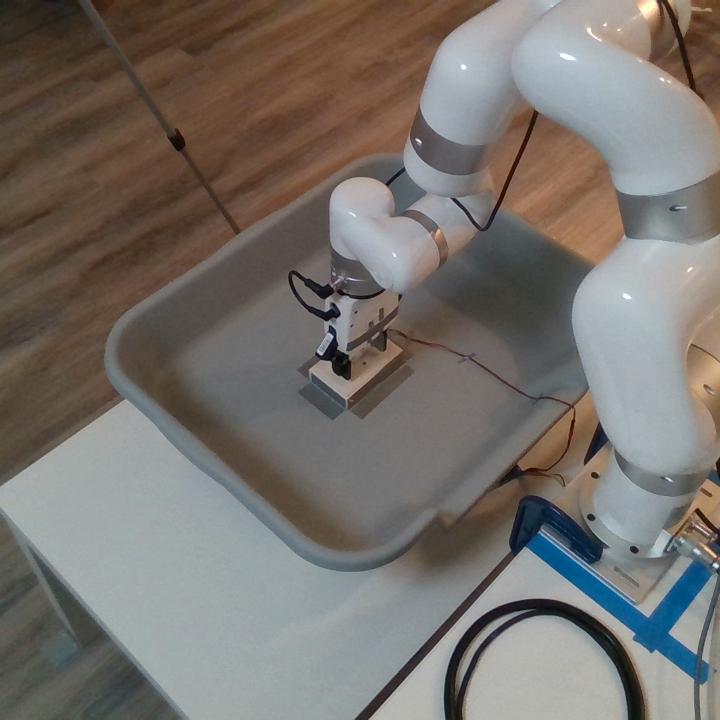}\\
        \includegraphics[width=\linewidth]{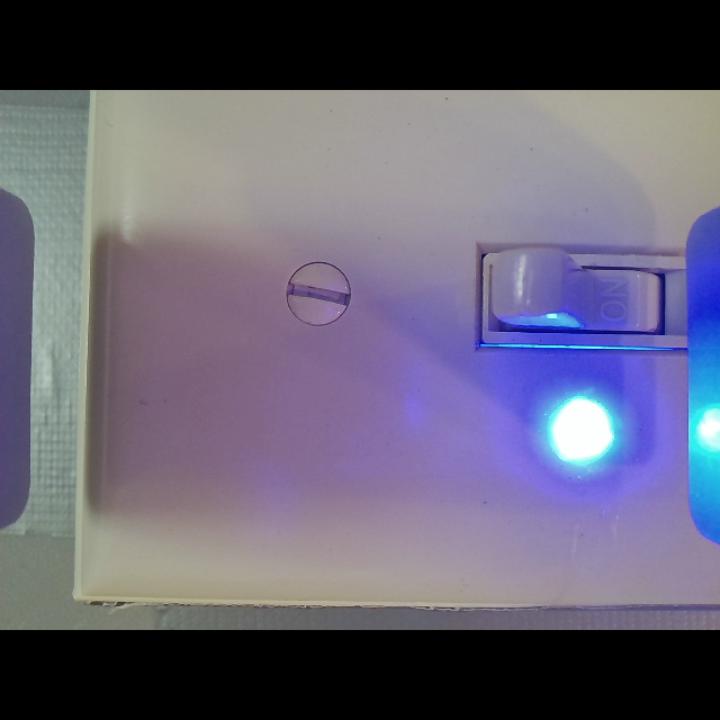}\\
        \includegraphics[width=\linewidth]{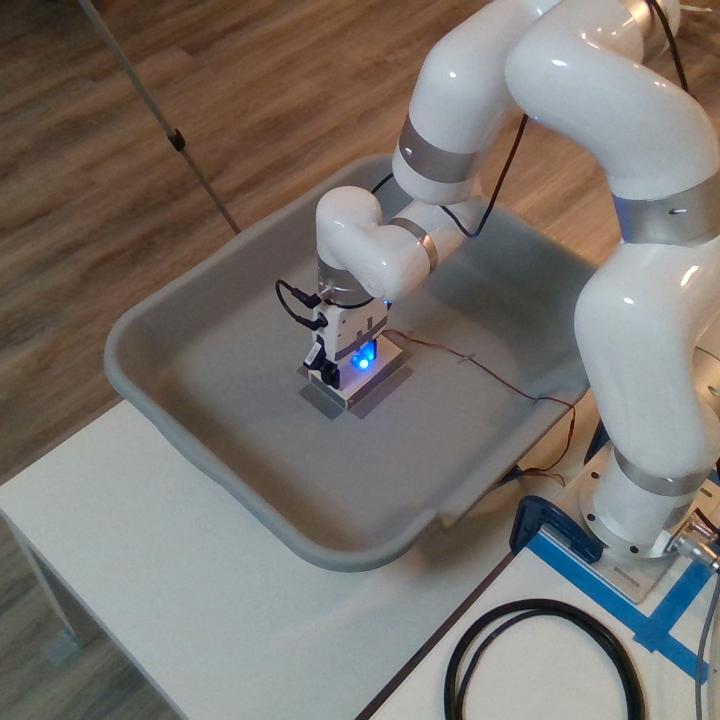}
      \end{subfigure}
      \caption{Light Switch}
    \end{subfigure}
    \hspace{.005\linewidth}
    \begin{subfigure}{\triplewidth\linewidth}
      \begin{subfigure}[c]{.64\linewidth}
        \centering
        \includegraphics[width=\linewidth, trim={56px 0px 112px 0px}, clip]{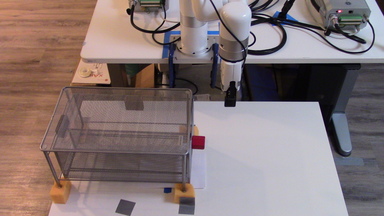}\\
        \includegraphics[width=\linewidth, trim={56px 0px 112px 0px}, clip]{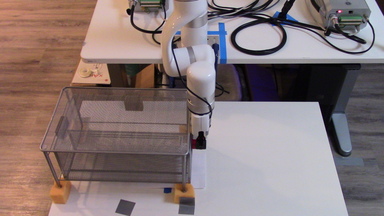}\\
        \includegraphics[width=\linewidth, trim={56px 0px 112px 0px}, clip]{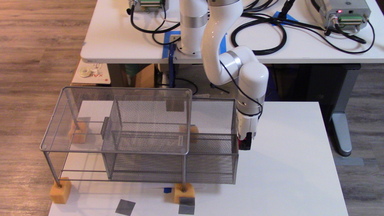}
      \end{subfigure}
      \begin{subfigure}[c]{.311\linewidth}
        \centering
        \includegraphics[width=\linewidth]{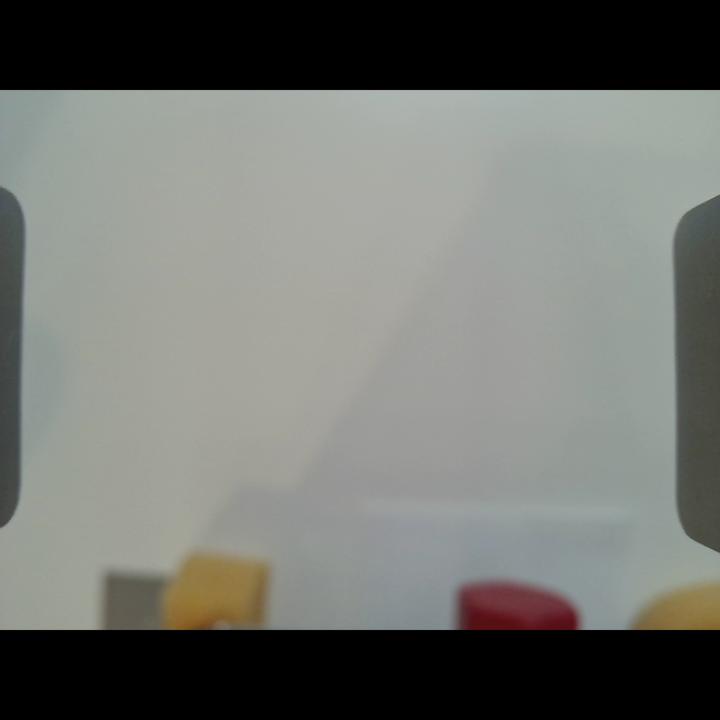}\\
        \includegraphics[width=\linewidth]{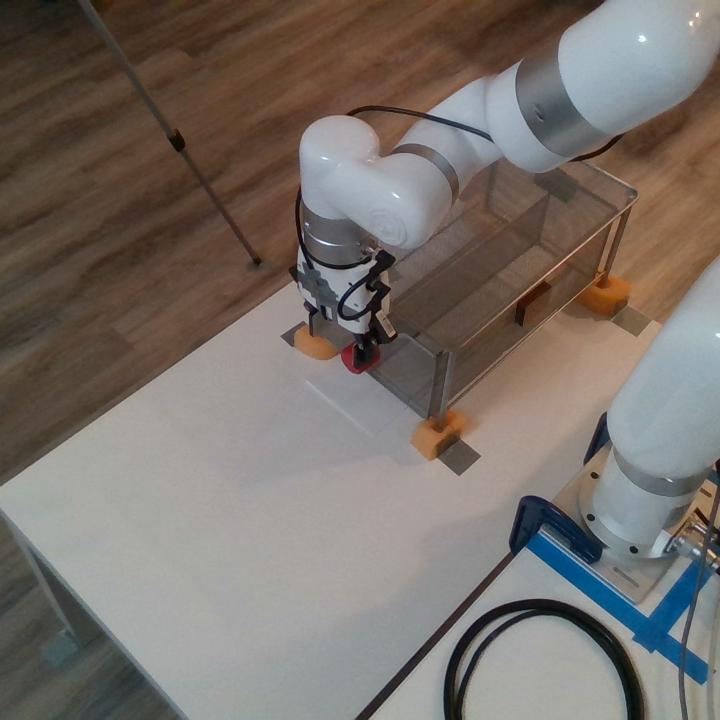}\\
        \includegraphics[width=\linewidth]{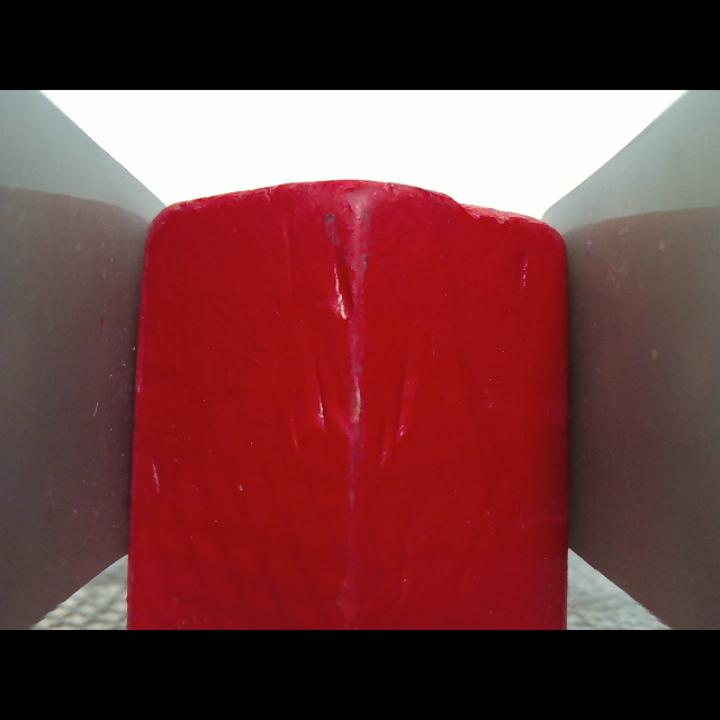}\\
        \includegraphics[width=\linewidth]{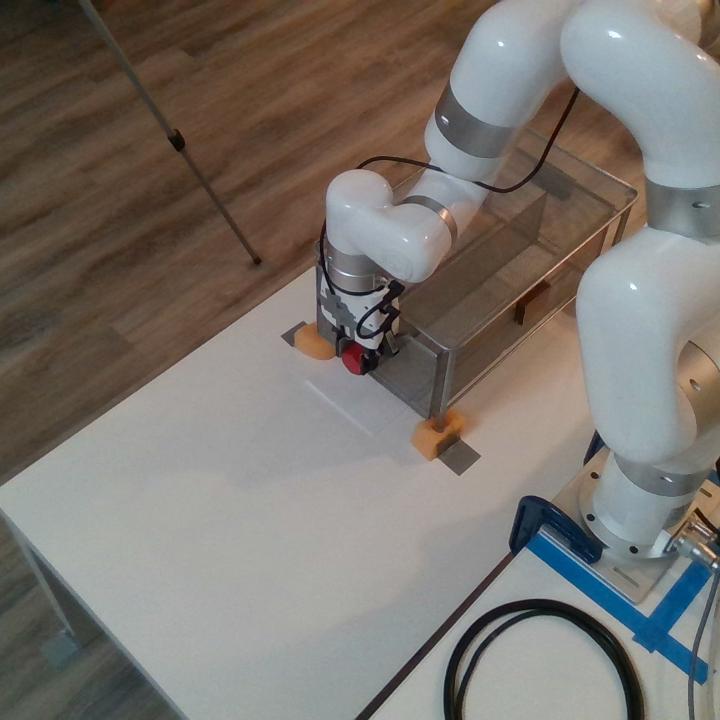}\\
        \includegraphics[width=\linewidth]{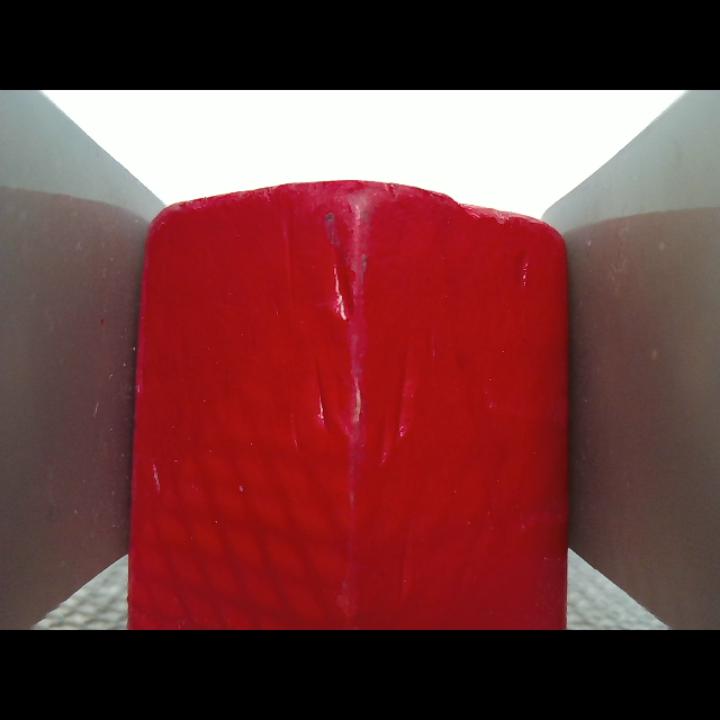}\\
        \includegraphics[width=\linewidth]{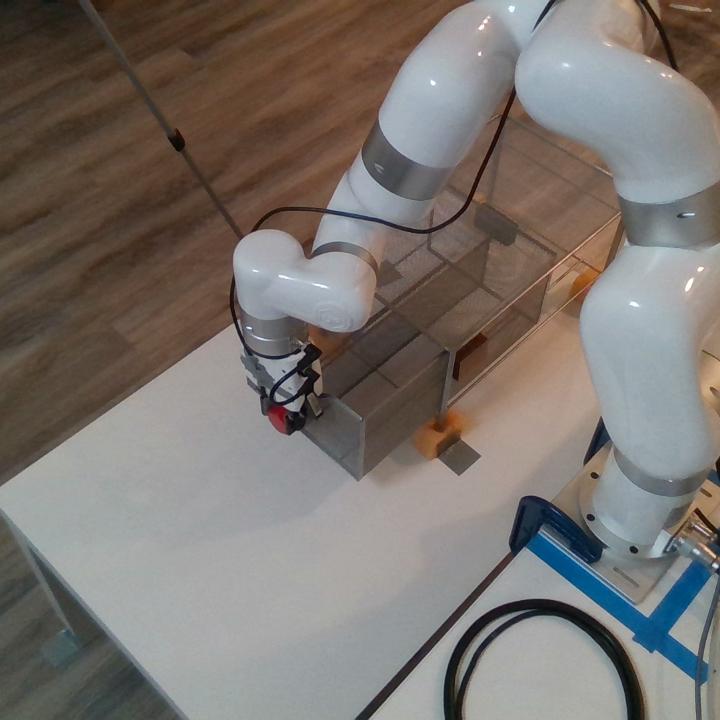}
      \end{subfigure}
      \caption{Drawer Open}
    \end{subfigure}
  \caption{The set of real world tasks used in this work, along with their pixel observations.
  Each column shows initial, intermediate, and completion states of a rollout during evaluation of our optimal policy.
  The right two images comprise the processed camera image input, which are concatenated and used as the observational input for the RL agent.
  %The right two images are the processed camera input, which are concatenated and used as input to determine the gripper position and aperture displacement.
  The sparse reward is only given when the robot completes the task.
  \ferm is able to solve all 6 tasks within an hour, using only 10 demonstrations.}
\label{fig:envs}
\vspace{-5mm}
\end{figure*}

%% file: figures/main_results_time.tex
\begin{figure*}[!ht!]
    \vspace{5pt}
    \includegraphics[width = \textwidth, trim={0px 0px 0px 35px}, clip]{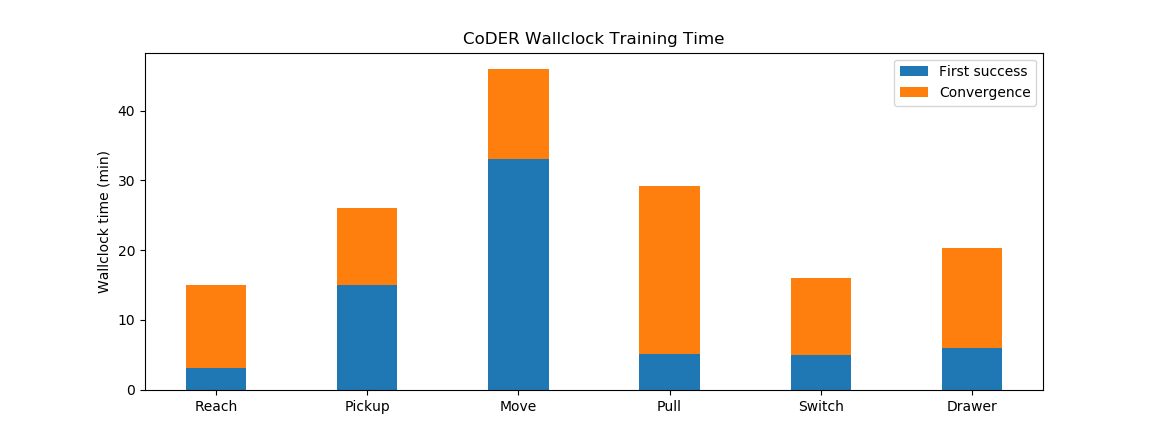} 
    % \fbox{\rule[-.5cm]{0cm}{4.5cm} \rule[-.5cm]{13cm}{0cm}}
  \caption{The speed at which our agents learn to complete the tasks.
Plotted above are the times at which the policy first achieves a success, as well as when an optimal policy is learnt. 
Our method starts to complete the tasks in around 30 minutes of training, and as little as 3 minutes for simple tasks such as Reach.}
\label{fig:res_time_bp}
\vspace{-6.5mm}
\end{figure*}

%% file: content/approach.tex
\section{Method}

Our proposed method, shown in Figure~\ref{fig:architecture}, combines demonstrations, unsupervised pre-training, and off-policy model-free RL with data augmentation into one holistic framework. \coder has three distinct steps -- (i) minimal collection of demonstrations (ii) encoder initialization with unsupervised pre-training and (iii) online policy learning through RL with augmented data
-- which we describe in detail below.

{\bf Minimal Collection of Demonstrations:} We initialize the replay buffer with a small number of expert demonstrations (we found $10$ to be sufficient) for each task. Demonstrations are collected with a joystick controller, shown in Figure~\ref{fig:teaser}. Our goal is to minimize the total time required to acquire a skill for an RL agent, including both policy training as well as time required to collect demonstrations. While collecting a larger number of demonstrations certainly improves training speed, we find 10 demonstrations is already sufficient to learn skills quickly (see Fig.~\ref{fig:abalation_demo} \textbf{Left}). For real world experiments, collecting $10$ expert demonstrations can be done within $10$ minutes
% (see Table~\ref{tab:robot_time_SR}), 
which includes the time needed to reset the environment after every demonstration. 

{\bf Unsupervised Encoder Pre-training:} After initializing the replay buffer with 10 demonstrations, we pre-train the convolutional encoder with instance-based contrastive learning, using stochastic random crop \cite{laskin2020curl} to generate query-key pairs. 
The key encoder is an exponentially moving average of the query encoder \cite{kaiming2019moco}, and the similarity measure between query-key pairs is the bi-linear inner product \cite{oord2018representation} shown in \ref{eq:infonce}. 
Note that the bi-linear inner product is only used to pre-train the encoder. 
After pre-training, the weight matrix in the bi-linear measure is discarded. 

{\bf Reinforcement Learning with Augmented Data:} After pre-training the convolutional encoder on offline demonstration data, we train a SAC \cite{haarnoja2018soft} agent with data augmentation \cite{laskin_lee2020rad} as the robot interacts with the environment. Since the replay buffer was initialized with demonstrations and SAC is an off-policy RL algorithm, during each minibatch update the agent receives a mix of demonstration observations and observations collected during training when performing gradient updates. The image augmentation used during training is random crop -- the same augmentation used during contrastive pre-training. 

%These three steps -- collecting 10 demonstrations, contrastive pre-training of the encoder, and off-policy RL with data augmentation -- comprise our framework for efficient robotic manipulation. \pz{Might be redundant.}

%% file: content/results.tex
\input{figures/main_results_barplot}

\section{Experimental Evaluation}

\input{content/robotic_setup.tex}

\input{content/environments}

\input{content/robotic_results_table_short}

\input{content/robotic_results}

\input{content/ablations.tex}

%In this section, we investigate the efficacy of our proposed method -- \coder. Our goal is to provide a simple yet effective baseline for robotic manipulation from pixels that is accessible to other researchers. Our hypothesis is that contrastive pre-training combined with data augmentated RL should result in data-efficient training given a handful of demonstrations to reduce the exploration challenge in the presence of sparse rewards. 

%Since \coder is composed of three independent ingredients, we ablate how each piece contributes to the overall framework. In addition to our hypothesis, we investigate the contribution of each component of the framework by answering the following questions:  (1) Are demonstrations required to learn efficiently and, if so, how many? (2) How does contrastive pre-training affect the performance of our agent and how many updates are required for initialization? (3) How important is data augmentation during online training of RL?

%% file: figures/main_results_barplot.tex
\begin{figure*}[!ht!]
    \vspace{0pt}
    \includegraphics[width = \textwidth]{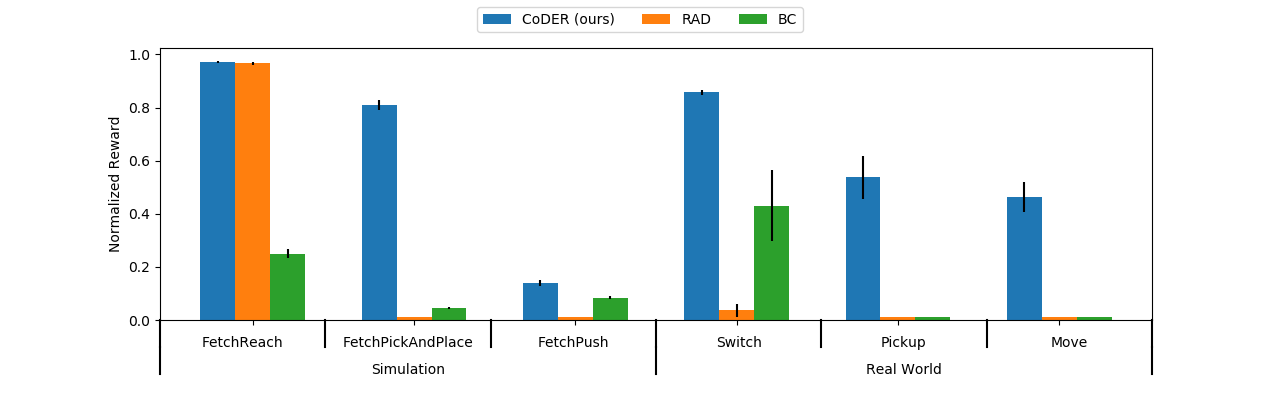} 
    % \fbox{\rule[-.5cm]{0cm}{4.5cm} \rule[-.5cm]{13cm}{0cm}}
  \caption{Baseline Comparisons. Shown are normalized rewards of the agent at the end of training for the simulated as well as the real robot results, as well as standard error. Reward is normalized by the maximum possible reward for all environments. While \coder is able to learn all the tasks, the baseline RL agent (RAD) is able to learn only one sparse reward task without demonstrations. Conversely, with access to only 10 demonstrations, behavior cloning is unable to learn in the more difficult environments, and only succeeds on the simpler tasks (FetchReach, Light Switch). \coder and RAD are trained for 200k environment steps in simulated tasks, and until convergence for real world tasks (30 episodes for Switch and Pickup, 60 episodes for Move). BC is trained over the dataset for 200 epochs for both simulated and real world tasks. Simulated tasks are evaluated over 100 episodes, while real world tasks are evaluated over 30 episodes.}
\label{fig:main_res_bp}
\label{fig:baselines}
\vspace{-6mm}
\end{figure*}

%% file: content/robotic_setup.tex
\subsection{Experimental Setup}

\textbf{Real robot}:
We use the xArm \cite{xarm} robot for all real-world experiments. The end effector, a parallel two-jaw gripper, is position controlled with three degrees of freedom.
At each step, the robot takes in an action containing the end effector and gripper aperture displacement.

\textbf{Operation space}:
The range of motion of the gripper is confined to a $25\,$cm-high imaginary box above the manipulation surface.
For majority of the tasks, objects are contained in a plastic tray approximately $40 \times 34\,$cm in size measured at its bottom. Sponge padding was placed below the tray to absorb minor collisions between the gripper and the objects.

\textbf{Input}: 
We use two RGB cameras, one positioned over the shoulder for maximal view of the arm, and the other located within the gripper to provide a local object-level view.
The over-the-shoulder camera is an Intel Realsense D415, with native resolution of $1280 \times 720$, and only the RGB frames are utilized during both training and testing.
Inside the gripper, we use an Arducam 8MP camera module configured to $640 \times 480$ in resolution.
Image frames from both cameras are cropped and down-sampled to $100 \times 100$ pixels for use in our training algorithm.

\textbf{Demonstrations}:
Using an Xbox controller \cite{xbox}, we teleoperate the robot by supplying the end effector and gripper aperture displacement.
Collecting demonstrations for each task requires less than \textbf{10 minutes}.
This is a very loose upper bound, which includes physically resetting the environment, and also accounts for possible human error and inefficiency during the demonstration collection phase.

%% file: content/environments.tex
\subsection{Environments and Baselines} 

{\bf Environments:} We evaluate \coder on six real-robotic manipulation tasks - reaching an object, picking up a block, moving a block to a target destination, pulling a large deformable object, flipping a switch, and opening a drawer. 
The block manipulation tasks (reach, pickup, move) are real-world adaptations of tasks from the OpenAI Gym Fetch suite \cite{openaigym}. We utilize these three OpenAI gym environments for simulated environment experiments. Since our method uses demonstrations, we include pull, which has been used in prior work on imitation learning \cite{rahmatizadeh2017vision, FlorenceMT20SSCVisuomotor2020}.  
Flipping a switch is included as it demands precision, while drawer opening is a common task in existing simulated robotic benchmarks
\cite{metaworld}. 
% Details of task setup are provided in the supplementary material.
% Section \ref{sec:task_details_short}. 

{\bf Baselines:} We compare \ferm to RAD~\cite{laskin_lee2020rad}, a leading supervised RL algorithm in simulated environments, and behavior cloning for our main results in Fig.~\ref{fig:main_res_bp}. In the ablations section, we investigate each individual component of \fermNoSpace. We investigate the contribution of each component of the \ferm algorithm by removing one component - demonstrations, contrastive pre-training, or data augmentation - while keeping others fixed.

%% file: content/robotic_results_table_short.tex
\begin{table*}[ht]
\begin{center}
\vspace{4mm}
\caption{
The success rates when evaluating the final policy learned by \coder over 30 episodes. Our method is able to achieve perfect success rate on the simpler tasks (Reach, Pickup, Light Switch, Drawer Open), and high success rates on the harder tasks (Move, Pull).}
\label{tab:robot_time_SR}
\begin{small}
\begin{sc}

\begin{tabular}{lccccccccc}
\toprule
Tasks & Reach &  Pickup & Move & Pull & {\begin{tabular}[c]{@{}c@{}} Light \\ Switch \end{tabular}}  & {\begin{tabular}[c]{@{}c@{}} Drawer \\ Open \end{tabular}}  \\
\midrule
\# Successes (/30)
& 30 & 30 & 26 & 28 & 30  & 30 \\ 
Success Rate (\%)  
& 100  & 100  & 86 & 93 & 100  & 100 
\\ 
\bottomrule
\end{tabular}
\end{sc}
\end{small}
\end{center}
% \vspace{-6mm}
\end{table*}

%% file: content/robotic_results.tex
\subsection{Results}
The main results of our investigation, including the time required to train an optimal policy as well the first successful task completion, are shown in Figure~\ref{fig:main_res_bp} and Table~\ref{tab:robot_time_SR}. We summarize the key findings below:

\vspace{0.2cm}

%\begin{center}
%1) \textit{What is the sample efficiency of our method, and is it general to a diverse set of tasks}.  

%\end{center}

\noindent (i) On average, \coder enables a single robotic arm to learn  optimal policies across all 6 tasks tested {\bf within 30 minutes of training time} with a range of 15-50 minutes, which corresponds to to 20-80 episodes of training. (see Figure~\ref{fig:res_time_bp} and Table~\ref{tab:robot_time_SR}).

\noindent (ii) When evaluated on 3 simulated and 3 real-robot tasks, \ferm substantially outperforms RAD and behavior cloning baselines (see Figure~\ref{fig:main_res_bp}).

\noindent (iii) The time to first successful task completion is {\bf on average 11 minutes} with a range of 3-33 minutes. The final policies achieve an {\bf average success rate of 96\%} with a range of  86-100\% across the tasks tested, suggesting that they have converged to near-optimal solutions to the tasks.

\noindent (iv) Collecting demonstrations and contrastive pre-training does not introduce significant overhead in terms of time. Collecting 10 expert demonstrations with a joystick requires 10 minutes of human operation. Contrastive pre-training completes within one minute on a single NVIDIA 2080Ti GPU. \\

\noindent (v) \coder solves all 6 tasks using the {\bf same hyperparameters} and without altering the camera setup, which demonstrates the ease of use and generality of the framework.

\vspace{0.2cm}

Altogether, an RL agent trained with \coder is able to learn optimal policies for the 6 tasks extremely efficiently. 
% While prior work in RL was able to solve dexterous manipulation tasks using RL with demonstrations in 2-3 hours of training \cite{zhu2018dext}, it also utilized dense rewards and more demonstrations. 
\coder is a reinforcement learning method that is able to solve a diverse set of sparse-reward robotic manipulation tasks directly from pixels in less than one hour.

%% file: content/ablations.tex
\subsection{Ablations}
\label{sec:ablations}

%\textbf{Sim environments:} We use the Fetch environments from the OpenAI Gym suite \cite{brockman2016openai} for our simulated ablation experiments.
%At each time-step, the xyz position of the gripper as well as the gripper aperture receive continuous-valued controls, same as our real robot setup.

%Within the Fetch suite, we mainly focus on the reach, push and pick-and-place tasks as they are similar to our real robot tasks. For visual distinctness, the goal location is marked as a red sphere, and the object is colored yellow.

%\begin{center}
%3) \textit{How important are the components of our framework?}
%\end{center}

\begin{figure*}[!htp]
    % \centering
    % \begin{subfigure}[c]{1.6cm}
    % \includegraphics[height=1.5cm]{figures/ablation_demo_cam1.png}\\
    % \includegraphics[height=1.5cm]{figures/ablation_demo_cam2.png}
    % \end{subfigure}
    \hspace{4mm}
    \begin{subfigure}[c]{5.2cm}
    \includegraphics[width=5.2cm, trim={10px 5px 40px 20px}, clip]{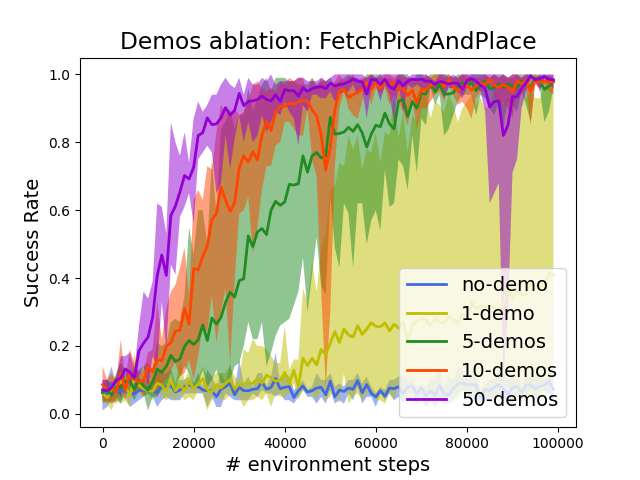}
    \end{subfigure}
    % \caption{
    % We ablate the number of demonstrations required by  FERM, and find that though the agent fails to learn with zero demonstrations, it can learn the pick-and-place task efficiently using only 10 demonstrations. }
    % \label{fig:abalation_demo}
    % \centering
    \begin{subfigure}[c]{5.4cm}
    \includegraphics[width=5.4cm, trim={0px 0px 40px 10px}, clip]{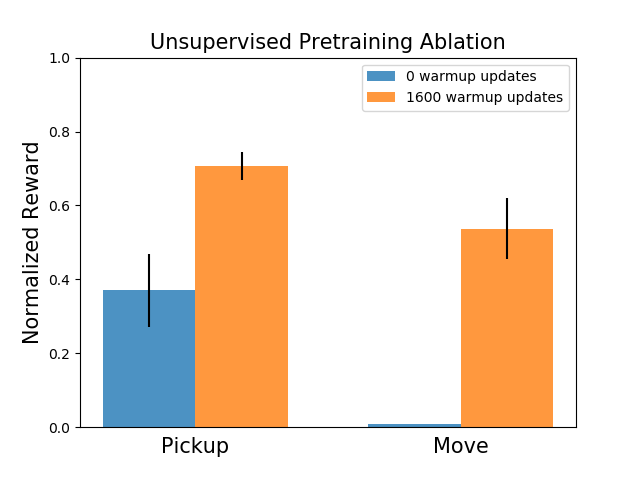}
    \end{subfigure}
    % \caption{
    % We compare the performance of the move task with and without the use of pre-training on the real xArm robot. The plotted episode returns during training show that the pick and move task fails to learn without contrastive pre-training. }
    % \label{fig:ablation_unsupervised}
    % \vspace{-6mm}
    % \centering
    % \raisebox{0.5cm}{
    \begin{subfigure}[c]{5.2cm}
    \includegraphics[height=4.3cm, trim={0px 5px 0px 10px}]{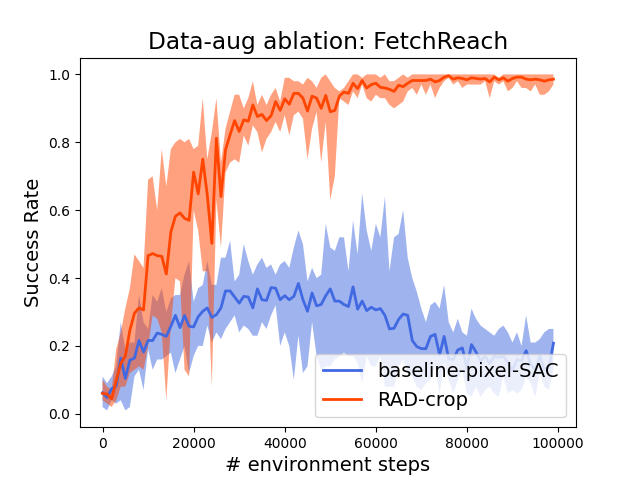}
    \end{subfigure}
    % }
    % \includegraphics[height=3.5cm, trim={10px 0px 40px 20px}, clip]{figures/ablation_data_aug_b.png}

    \caption{
    \textbf{Left:} We ablate the number of demonstrations required by \fermNoSpace, and find that although the agent fails to learn with zero demonstrations, it can learn the PickAndPlace task efficiently using only 10 demonstrations.
    \textbf{Center:} We compare the performance of the move task with and without the use of pre-training on the real xArm robot. The plotted episode returns at convergence show that the contrastive pre-training substantially boosts performance. 
    \textbf{Right:} 
    % A single camera view is provided as the observation (left). 
    Using data augmentation, the agent achieves successful performance. Using non-augmented observations, the agent fails to learn the task. 
    % \ml{ Much better now that these have all been merged! Small edits: left fig overlaps with center. Center fig should be bar plot with confidence intervals.}
    }
    \label{fig:abalation_demo}
    \label{fig:ablation_unsupervised}
    \label{fig:abalation_data_aug}
    %\vspace{-6mm}
\end{figure*}

\begin{figure}[ht!]%{r}{0.65\textwidth}
    \centering
  \begin{subfigure}[t]{.30\linewidth}
    \centering\includegraphics[width=.9\linewidth]{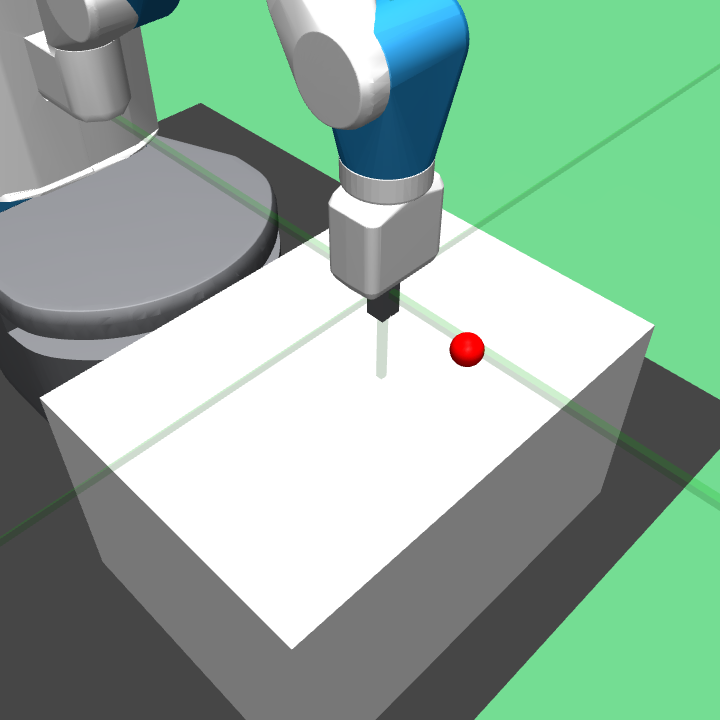}
    \caption{FetchReach}
  \end{subfigure}
  \begin{subfigure}[t]{.35\linewidth}
    \centering\includegraphics[width=0.77139\linewidth]{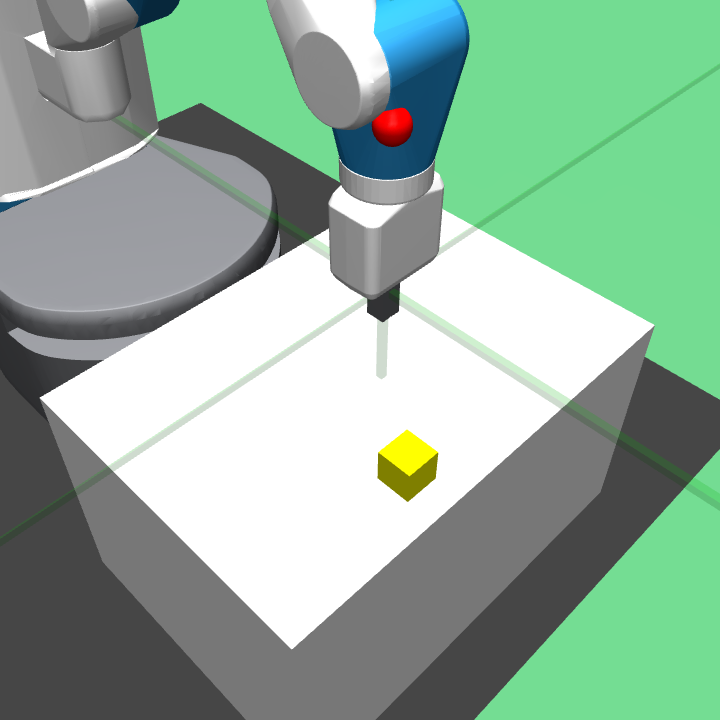}
    \caption{FetchPickAndPlace}
  \end{subfigure}
  \begin{subfigure}[t]{.30\linewidth}
    \centering\includegraphics[width=.9\linewidth]{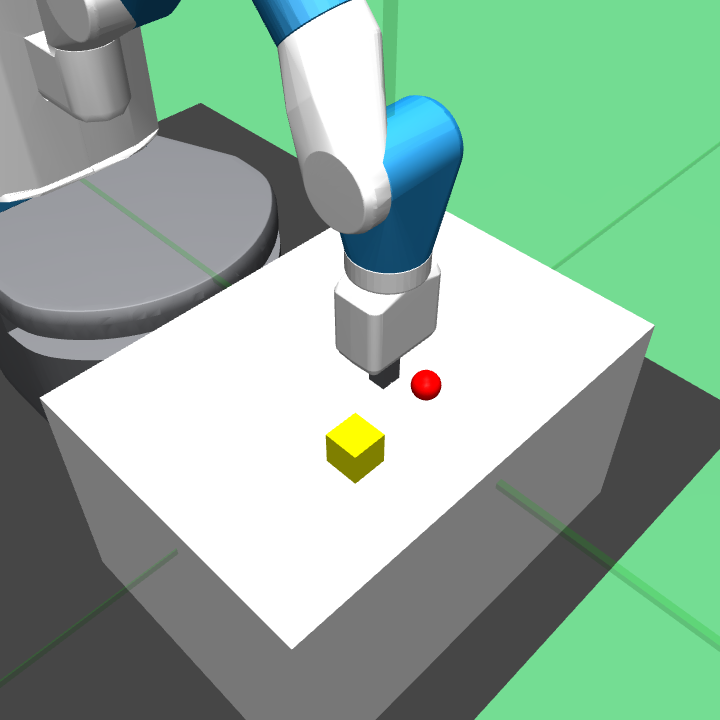}
    \caption{FetchPush}
  \end{subfigure}
  \caption{
  OpenAI gym~\cite{openaigym} environments used in addition to real-robot experiments. 
%   Shown are sample observations from the third person camera, as well as 
%   \textbf{FetchReach}: The gripper must move to the target location, shown in red. \textbf{FetchPickAndPlace}: The gripper must pick the object, and place it on the goal. \textbf{Right}: FetchPush. The gripper is used to push the object to the goal.
  %In addition to the real robot experiments, we also include simulated experiments to investigate the core components of the FERM method. We utilize the Fetch Gym suite of environments \cite{openaigym}, which includes the reach, push, and pick-and-place tasks.
  }
  \label{fig:fetch_envs}
  \vspace{-3mm}
\end{figure}

In this section, we investigate  how the three core components of \coder -- demonstrations, contrastive pre-training, and data augmentation -- contribute to the overall efficiency of the framework through ablations in simulated environments, shown in Figure~\ref{fig:fetch_envs}. 
%For additional ablations, we refer the reader to Section~\ref{sec:further_ablations}.

{\it How many demonstrations are needed?} 
% In real robot settings, assigning dense rewards is often difficult or infeasible.

While sparse rewards are simpler to define, they pose an exploration challenge since the robot is unlikely to randomly stumble on a reward state.
We address this issue by providing demonstrations to the RL agent. We ablate the number of demonstrations required to learn efficiently on the simulated pick and place task in Figure~\ref{fig:abalation_demo}. 
We find that while the agent fails entirely with zero demonstrations, it is able to start learning the task with just one demonstration. 
While more demonstrations improve learning efficiency and reduce the variance of the policy, ten demonstrations suffice to learn quickly.
We then evaluate the effectiveness of the 10 demonstrations by comparing our method to training behavior cloning. As shown in Figure~\ref{fig:main_res_bp}, the 10 demonstrations are not enough to learn an effective policy. 
% Refer to the supplementary material for further details.

%However, when using a sparse reward, it is incredibly difficult to interact with the reward if the task is difficult. \AZ{Should we include a figure for like lots of steps on the different tasks with sparse reward, RAD}
%We address this issue by supplying 10 human demonstrations to expose the agent to sparse reward signals that would otherwise be hard to obtain through random exploration. 
%While only supplying sparse reward, we ablate on the number of demonstration provided to the agent to determine the best number of demonstrations used in real world applications. 
%Our result in Figure \ref{fig:abalation_demo} shows that as little as 1 single demonstration enables the policy to explore and complete the task, while using more demonstrations further boosts training and reduces variance across different seeds.

{\it How important is unsupervised contrastive pre-training?} We next study the role of contrastive pre-training in \coder. We ablate our method with and without contrastive pre-training on the real world pickup and move task, shown in Figure~\ref{fig:ablation_unsupervised}, where we compare with $0$ and with $1600$ iterations of pre-training to initialize the encoder.
%While no pre-training fails to learn a policy, the pre-trained variants are able to solve the task, with the $1600$ iteration variant 
%in real world experiments, we found the pre-training to be crucial for fast and efficient learning.
%Shown in Figure \ref{fig:ablation_unsupervised}, we train the real robot to perform the Move task using $0$, $100$, and $1600$ iterations of pre-training.
With $1600$ contrastive iterations, the agent is able to learn a successful policy while the other runs fail to learn. In the case of no pre-training, the agent is only able to succeed once during the entire hour of training.

{\it Is online data augmentation necessary?} To justify the use of data augmentation during online RL training, we compare the performance of SAC with and without data augmentation for a simple, dense reward reaching task.
In the FetchReach environment, we use the dense reward $r = -d$ where $d$ is the Euclidean distance between the gripper and the goal.
As shown in Figure \ref{fig:abalation_data_aug}, without data augmentation, the RL agent is unable to learn the simple task, and asymptotically collapses, as shown by the evaluation success rate.
This motivates us to use data augmentation for our sparse reward tasks, which encounter even less learning signal. 
% to learn features.

% \begin{figure}[h!]
%     \centering
%     \includegraphics[height=3.5cm, trim={10px 0px 40px 20px}, clip]{figures/ablation_4_temp.png}
%     \caption{(Image being a temporary placeholder) Performance comparison for the Gym Fetch Pick-and-place task given different number of contrastive objective training steps.}
%     \label{fig:abalation_cpc}
% \end{figure}

%% file: content/related_work.tex
\section{Related Work}

%In what follows we discuss related work in imitation learning, reinforcement learning, data augmentation and unsupervised representation learning. \pa{can cut this preceding sentence if looking to save space; it's obvious from the headers anyway}

{\bf Imitation Learning:} Imitation learning is a framework for learning autonomous skills from demonstrations. 
One of the simplest and perhaps most widely used forms of imitation learning is behavior cloning (BC) where an agent learns a skill by regressing onto demonstration data. 
BC has been successfully applied across diverse modalities including video games \cite{DBLP:journals/jmlr/RossGB11}, autonomous navigation \cite{Pomerleau88alvinn,BojarskiTDFFGJM16}, autonomous aviation \cite{GiustiGCHRFFFSC16}, locomotion \cite{NakanishiMECSK04,DBLP:conf/iros/KalakrishnanBPS09}, and manipulation \cite{zhang2018DeepIL,young2020, rahmatizadeh2017vision}. Other imitation learning approaches include Dataset Aggregation \cite{dagger2010}, and Inverse Reinforcement Learning \cite{DBLP:conf/icml/NgR00,DBLP:conf/icml/PieterN04}. 
A general limitation of imitation learning approaches is the requirement for a large number of demonstrations for each task~\cite{sharma2018multiple}. Although recent advancements have shown that imitation learning can learn with a much more modest amount of demonstrations \cite{zhang2018DeepIL, rahmatizadeh2017vision, FlorenceMT20SSCVisuomotor2020}, \ferm can learn in the same number of episodes, of which the majority are spent with reinforcement learning.

{\bf Reinforcement Learning:} Reinforcement Learning (RL) has been a promising approach for robotic manipulation due to its ability to learn skills autonomously.
% , but has not achieved widespread adoption in real-world robotics. 
Recently, deep RL methods excelled at playing  video games from pixels \cite{mnih2015human,openai2019dota} as well as learning robotic manipulation policies from visual input \cite{levine2015end, finn2017deep,haarnoja2018soft}. However, widespread adoption of RL in real-world robotics has been bottle-necked due to the data-inefficiency of the method, among other factors such as safety. Though there exist prior frameworks for efficient position controlled robotic manipulation \cite{zhu2018dext, c2farm}, they still require hours of training while providing additional information such as a dense reward function, or heavy computation along with depth information. 
\coder is most closely related to other methods that use RL with demonstrations. 
Prior methods \cite{nair18rlwithdemos, rajeswaran2017learning, Vecerik2017leveraging} solve robotic manipulation tasks from coordinate state input, rather than image input, by initializing the replay buffer of an RL algorithm with demonstrations to overcome the exploration problem in the sparse reward setting. 
Recent advances in vision-based imitation learning is also able to learn with limited amount of demonstrations \cite{florence2022implicit, selfsupervisedcorrespondance, transporter}, however, the learned policy distribution is still limited by the distribution provided by the experts.

{\bf Data Augmentation:} Image augmentation refers to stochastically altering images through transformations such as cropping, rotating, or color-jittering. It is widely used in computer vision architectures including seminal works such as LeNet \cite{lecun1998} and AlexNet \cite{krizhevskySH17}. Data augmentation has played a crucial role in unsupervised representation learning in computer vision \cite{henaff2019data,kaiming2019moco,chen2020simclr}, while other works investigated automatic generation of data augmentation strategies \cite{CubukZMVL19}. Data augmentation has also been utilized in prior real robot RL methods \cite{kalashnikov2018qt}; however, the extent of its significance for efficient training was not fully understood until recent works \cite{laskin2020curl,laskin_lee2020rad,kostrikov2020image}, which showed that carefully implemented data augmentation makes RL policies from pixels as efficient as those from coordinate state. Finally, data augmentation has also been shown to improve performance in imitation learning \cite{young2020}. In this work, data augmentation comprises one of three components of a general framework for efficient learning. 

%Data augmentation has certainly been utilized in prior real robot RL methods \cite{kalashnikov2018qt}, however
%While data augmentation has been previously observed to succeed in the context of real robot RL within QT-OPT \cite{kalashnikov2018qt}, our methods differ in two significant factors.
%First, QT-OPT required a robot farm and over 800 robot hours to learn effective grasp policies, due to learning a generalizing grasp policy.
%Second, we supply a handful of demonstrations to tackle the sparse reward problem, while QT-OPT uses a policy that can successfully solve the task to provide exploration signal.
%Finally, data augmentation has also been shown to improve performance in imitation learning \cite{young2020}. 

{\bf Unsupervised Representation Learning:} The goal of unsupervised representation learning is to extract representations of high-dimensional unlabeled data that can then be used to learn downstream tasks efficiently.  Most relevant to our work is contrastive learning, which is a framework for learning effective representations that satisfy similarity constraints between a pair of points in dataset. In contrastive learning, latent embeddings are learned by minimizing the latent distance between similar data points and maximizing them between dissimilar ones. Recently, a number of contrastive learning methods \cite{henaff2019data,he2019momentum,chen2020simclr} have achieved state-of-the-art label-efficient training in computer vision. A number of recent investigations in robotics have leveraged contrastive losses to learn viewpoint invariant representations from videos \cite{Sermanet18tcn}, and learn object representations \cite{pirk2020}. In this work, we focus on instance-based contrastive learning \cite{wu2018unsupervised} similar to how it is used in vision \cite{kaiming2019moco,chen2020simclr}
 and RL on simulated benchmarks \cite{laskin2020curl, stooke2020atc}.

%% file: content/limitations.tex
\section{Conclusion and Limitations}
Although \coder enables data-efficient deployment of RL onto real robots, the method also has a number of limitations. 
First, like most RL algorithms, \coder may require assistance for resets, and \coder policies can only solve the tasks that they were trained on and while they may display some degree of generalization to small changes such as object shape or perturbations, we do not expect \coder policies to generalize to qualitatively different tasks that were unseen during training. 
Second, while the tasks considered in this paper are standard robotics evaluation tasks, they all have relatively short horizons. 
Since \coder relies on a sparse reward signal to learn, we do not expect this framework to succeed in long-horizon sparse reward tasks, where random interaction with the reward is unlikely. 
Finally, we expect the performance of \coder to degrade if the visual conditions of the scene change substantially, which is likely in non-lab settings with frequent background distractors and lighting changes. 
Rather than addressing generalization to new tasks and visual settings or long-horizon settings, this paper focuses on the data-efficiency problem of training RL policies on real robots. We believe that data-efficient generalization and long-horizon problem solving are important open problem in robot learning that we leave for future work. 

%% file: content/acknowledgements.tex
\section{Acknowledgements}
We gratefully acknowledge support from Open Philanthropy, Darpa LwLL, Berkeley Deep Drive and Amazon Web Services. We would also like to thank the reviewers for detailed feedback on our submission.